**Dynamic Silos: Increased Modularity in Intra-organizational Communication Networks During the Covid-19 Pandemic***

**Tiona Zuzul**
Harvard Business School, Boston, MA 02163, tzuzul@hbs.edu

**Emily Cox Pahnke**
Foster School of Business, University of Washington, Seattle, WA 98195, eacox@uw.edu

**Jonathan Larson, Patrick Bourke, Nicholas Caurvina**
Microsoft Research, Redmond, WA, 98052, {jolarso, pbourke, nick.caurvina}@microsoft.com

**Neha Parikh Shah, Fereshteh Amini, Jeffrey Weston**
Microsoft, Redmond, WA, 98052, {neha.shah, fereshteh.amini, jeffrey.weston}@microsoft.com

**Youngser Park, Joshua Vogelstein**
Johns Hopkins University, Baltimore, MD, 21218, {youngser, jovo}@jhu.edu

**Christopher White**
Microsoft Research, Redmond, WA, 98052, chwh@microsoft.com

**Carey E. Priebe**
Johns Hopkins University, Baltimore, MD, 21218, cep@jhu.edu

**ABSTRACT**

Workplace communications around the world were drastically altered by Covid-19, related work-from-home orders, and the rise of remote work. To understand these shifts, we analyzed aggregated, anonymized metadata from over 360 billion emails within 4,361 organizations worldwide. By comparing month-to-month and year-over-year metrics, we examined changes in network community structures over 24 months before and after Covid-19. We also examined shifts across multiple communication media (email, instant messages, video calls, and calendaring software) within a single global organization, and compared them to communications shifts that were driven by changes in formal organizational structure. We found that, in 2020, organizations around the world became more siloed than in 2019, evidenced by increased modularity. This shift was concurrent with decreased stability within silos. Collectively, our analyses indicate that following the onset of Covid-19, employees began to shift more dynamically between subcommunities (teams, workgroups or functional areas). At the same time, once in a subcommunity, they limited their communication to other members of that community. We term these network changes *dynamic silos*. We provide initial insights into the meaning and implications of dynamic silos for the future of work.

## INTRODUCTION

How has Covid-19 altered workplace communication? Mapping intra-organizational communication networks—who communicates with whom—is critical to understanding how work gets done in organizations (Blau 1963, Jacob and Watts 2021, Kleinbaum, Stuart, and Tushman 2013). Intra-organizational communication networks vary in their community structures—that is, the natural partitioning of individuals into strongly connected clusters or subcommunities (Aldecoa and Marín 2011, Fortunato 2010, Fortunato and Newman 2022, Newman and Girvan 2004). Covid-19 created exceptional circumstances—and a natural experiment—that disrupted work in many ways never observed on such a scale. The public health crisis, related work-from-home orders, and the subsequent rise of remote work redefined how employees could communicate, and digital forms of communication (e.g., email, instant messaging, video calls) began to supplant in-person social exchanges (DeFilippis et al. 2022, Gibbs et al. 2021). Did these changes reshape the community structures within intra-organizational communication networks? If so, how?

The extraordinary scale and scope of Covid-related work disruptions mean there is little prior theory to guide this question. The goal of this exploratory study is therefore not to test hypotheses, but to uncover broad patterns in the impact of Covid-19 on intra-organizational communication networks. Understanding how patterns of communication shifted at the community level is important because work in organizations is typically organized through collaboration within subcommunities such as teams, workgroups, and functional areas (Cummings 2004, Edmonson 1999, Gersick and Hackman 1990). On one hand, remote work and the rise of digital communication tools can provide flexibility and expand the boundaries of who can communicate with whom (e.g., Kellogg, Orlikowski, and Yates 2006), potentially creating links across subcommunities. On the other hand, the absence of in-person encounters might narrow the scope of employees' communication to essential connections exclusively within their subcommunity (e.g., Yang et al. 2022). Either shift might have important consequences for organizational performance and innovation.

To study shifts in intra-organizational communication networks, we calculated community-level measures from data on 362 billion aggregated email receipts between more than 1.4 billion email accounts in 4,361 organizations around the world. We constructed two measures of community structure commonly used by network scholars in computer science, epidemiology, neurology, and biology (e.g., Fortunato and Newman 2022, Girvan and Newman 2002, Kernighan and Lin 1970, Moore and Newman 2000, Jeong, Tombor, Albert, Oltavi and Barabási 2000): the Adjusted Rand Index (ARI) to measure stability (that is, how much membership within subcommunities changes), and modularity to measure siloing (that is, how much members communicate within and between subcommunities). We found that the volume of emails increased across organizations following Covid-related work-from-home orders. At the same time, a monthly year-over-year comparison showed that, while ARI decreased, modularity increased across organizations. Decreased ARI suggests that, following Covid, employees moved more frequently between teams or workgroups than in the past; that is, employees' membership within organizational subcommunities began to change more dynamically. Increased modularity suggests that, following Covid, employees began communicating more intensively with those inside their team or work group, and less intensively with those outside the group; that is, subcommunities became more siloed in their communication patterns. We conceptualize these shifts as *dynamic silos.*

To further examine these shifts, we analyzed additional data on multiple communication modalities within a single organization—Microsoft. We calculated community-level measures for 524 million email receipts, 2 billion instant messages, 70 million video calls, and 39 million scheduled calendar meetings from the entire organization over 24 months. We found that, following a firm-wide Covid-19 work-from-home order in March 2020, the volume of instant messages and video calls increased dramatically, suggesting that they were replacing in-person interactions, while the volume of emails and scheduled calendar meetings remained relatively stable, suggesting that these were not substituting for in-person interactions. We also find that ARI decreased and modularity increased across communication media—regardless of the presence or absence of apparent substitution—indicating that the observed changes revealed fundamental shifts in community structures. Once work moved online, organizational subcommunities became more siloed, even in the face of increased membership instability within the subcommunities.

Our findings provide a baseline set of results that can inform future research on the long-term implications of Covid-19, the rise of remote work, and the measurement of community structures within organizations. We extend recent studies examining organizational changes following the onset of Covid-19 (e.g., DeFilippis et al. 2022, Gibbs et al. 2021, Yang et al. 2022) in three important ways. First, most studies evaluate changes in a single organization; for example, Yang et al. (2022) examined communication patterns within Microsoft from December 2019 to June 2020. As DeFilippis et al. (2022: 2) recently wrote, "there is scarce prior evidence on digital communication across many firms, even in the absence of a crisis." Second, most studies observe changes over a relatively short period before and after Covid-19 restrictions; for example, DeFilippis et al. (2022) examined changes across organizations in the 16 weeks surrounding lockdown measures. As Yang et al. (2022: 51) wrote in response, "[I]t is possible that the long-term effects of firm-wide remote work are different" than these findings imply. In contrast, our main results were replicated both around the world and within one organization over 24 months. By conducting a monthly year-over-year comparison of communication patterns, we control for seasonal variation and highlight the long-term changes stemming from the shift to remote work.

Finally, this is the first study we know of to examine post-Covid shifts in community-level structures within organizational networks. Studies have examined individual-level effects associated with post-Covid remote work, finding that employees had longer work hours and fewer meetings (DeFilippis et al. 2022), fewer one-on-one meetings (Gibbs et al. 2021), and fewer individuals with whom they communicated (Yang et al. 2022). But subcommunities like teams and workgroups are at the heart of how work is done in organizations. And while scholars have theorized that Covid-19 may have reshaped organizational communities, including levels of "organizational modularization" (Foss, 2021: 272), research on shifts in community structures has been scant. As Foss (2021: 272) recently argued, "the problem with such research is that panel datasets with sufficient data on organizational structure or communication to allow for comparison across organizations and over time "are basically non-existing."

We thus leverage our uniquely large data set to examine how Covid-19 and the shift to remote work were associated with changes in community structures within organizations.

More generally, management studies have shown that the structural properties of organizational networks—for example, structural holes and cohesion—are linked to innovation, learning, and performance (Obstfeld 2005, Paruchuri and Awate 2017, Reagans and McEvily 2003, Soda, Tortoriello, and Iorio 2018). But while research in diverse disciplines (from computer science to biology) has demonstrated community structures to be among the most important properties of networks (e.g., Fortunato and Newman 2022, Girvan and Newman 2002, Kernighan and Lin 1970, Moore and Newman 2000, Jeong, Tombor, Albert, Oltavi and Barabási 2000), few studies have examined community structures within intra-organizational networks. More broadly, Jacobsen et al.'s (2022) review of network research in the social sciences noted only four studies at the intra-organizational level of analysis. We attribute this to computational and data limitations. Advances in computational capabilities have only recently enabled the large-scale study of email networks and the measurement of their community structures. Detection of community structures is also only possible within large networks (Fortunato 2010, Fortunato and Newman 2022) and access to sufficiently large intra-organizational datasets poses a challenge. We thus leverage new computational capabilities and our data, providing unprecedented insight into the community structures of intra-organizational networks both around the world and within a single organization, to generate new ideas about how Covid-19 affected intra-organizational communication. To aid future researchers, we also contribute by making available a generative model to reduce computational intensity and by providing access to part of our dataset.

The remainder of this paper is organized as follows. We anchor our study in research on intra-organizational networks and organizational design. We then describe the data and our methods of analysis. We present results at two levels (across organizations and within Microsoft) and then describe the generative model we created to facilitate future research. Finally, we highlight the implications of our results to inspire future research on intra-organizational networks, community structures, and the impact of Covid-19 on organizational performance and innovation.

**Understanding Intra-organizational Communication Networks**

Scholars have long recognized that formal organizational charts may not accurately represent actual communication patterns in the workplace. Blau (1963) highlighted the importance of informal "water cooler" conversations versus formal organizational structures in understanding how employees work. Dalton (1959) and Strauss (2018) explored how formal organizational structures could be subverted by employees, affecting output and efficiency. More recent research has examined how employee networks may differ even within the same organization. For example, women tend to have more network ties than men (Kleinbaum, Stuart, and Tushman 2013), but gain fewer advantages from those ties (Ibarra1992), and are less frequently introduced to new, potentially valuable ties (Abraham 2020). The structural properties of employees' networks can influence outcomes including innovation and learning (Paruchuri and Awate 2017, Reagans and McEvily 2003). For example, Paruchuri and Awate (2017) showed that inventors who spanned more structural holes within semi-conductor firms were more likely to engage in local search when filing new patents. Reagans and McEvily (2003) found that increased network range and cohesion facilitated knowledge transfer between employees at a contract manufacturing firm. Examining the structure of actual—rather than assigned—networks is critical for understanding how work unfolds in an organization.

Email is one way to measure communication networks within organizations **(**Jacobs and Watts 2021, Kleinbaum et al. 2013). By examining patterns of email, it is possible to observe social networks that aggregate patterns of individual interaction to the organization level. These aggregate patterns can provide insights into how information flows in an organization and how flows vary between organizations. For example, Jacobs and Watts's (2021) exploratory study found that more geographically dispersed organizations had more centralized email networks, indicating that longer paths were required to access information. While email receipts do not capture all possible interpersonal interactions,[1] scholars have argued that email data "is a representation of the actual flows of communication" (Quintane

[1] As detailed later, we also consider alternate modes of communication to construct intra-organizational networks.

and Kleinbaum 2011) and "may well [make up] the majority of interactions in an organization" (Kleinbaum, Stuart, and Tushman 2013) where work is done in offices and requires frequent written communication.

Constructing social networks based on email data can help identify higher-level community structures within organizations. Community structures reveal the structure of ties within a network in order to partition nodes into densely connected clusters (Aldecoa and Marín 2011, Fortunato 2010, Fortunato and Newman 2022, Newman and Girvan 2004). For example, one well-known type of community structure is a small world (Watts and Strogatz 1998), in which the network, or a portion of it, is characterized by dense clustering of ties and short path lengths between nodes, facilitating rapid and efficient information flow (Gulati et al. 2012). Plentiful research has applied ideas from statistical physics and mathematics to infer and demonstrate the importance of community structures in, among others, epidemiological, metabolic, neural, biological, and computational networks (e.g., Girvan and Newman 2002, Kernighan and Lin 1970, Moore and Newman 2000, Oltavi and Barabási 2000). But while scholars since the 1940s (e.g., Davis, Gardner, and Gardner 1941) have argued that individuals' subgroups should be identified from patterns of interaction or communication rather than from exogenously determined categories (cf. Katz et al. 2004), studies measuring community structures within organizations are scant. We leverage recent advances in computational capabilities and unprecedented data on intra-organizational communication patterns within multiple organizations to construct such structures and examine how they shifted following Covid-19.

**Examining Shifts in Community Structures following Covid-19**

By examining community structure, we extend recent studies examining the individual-level outcomes associated with the post-Covid-19 shift to remote work. Several studies found that employees significantly increased their average hours worked (DeFilippis et al. 2022, Gibbs et al. 2021, Kun et al. 2022). Scholars have also noted shifts in whom individuals communicated with and how communication occurred. For example, DeFilippis et al. (2022) found that employees across geographies spent less time in meetings, although they held more meetings with a larger number of attendees. Yang et al. (2022)

found that the shift to remote work lowered employees' number of bridging ties and reduced the time they spent communicating with weak ties. Similarly, Gibbs et al. (2021) showed that employees had fewer contacts both inside and outside their organization. While these findings have implications for community-level measures, research has not explored shifts in such structures directly.

We have several reasons for extending early insights into the impact of Covid-19 on intra-organizational communication networks to examine shifts in community structures. First, this does not require imposing a formal organizational structure; community structures can be inferred. Inferring rather than imposing structure allows for an understanding of how communication flows in practice between individuals, teams, and functions (Katz et al. 2004). Second, we expect shifts in community structure to produce macro-level changes in information and knowledge flow (see, for example, Gulati et al. 2012) with potentially long-lasting impact on organizational performance and innovation.

We focus on two measures of a network's community structure: ARI and modularity. ARI measures the stability of subcommunity membership: that is, the movement of members in and out of a subcommunity over time. While traditional research in organizations has examined subcommunities (for example, teams) with stable membership (e.g. Gersick 1988), recent work points to "rapidly shifting membership" or "temporal instability" as "new dimensions of teamwork" (Kerrissey, Satterstrom and Edmondson 2020: 2-4). Modularity captures the degree to which subcommunities are siloed: that is, the extent to which members of a subcommunity communicate with each other as opposed to members of other subcommunities. Highly modular networks are one form of small world (Watts and Strogatz 1998); networks with low modularity have less–well-defined silos and greater overall connectivity.

We conceptualize ARI and modularity as two distinct dimensions of community structure (see Table 1 for a mapping of these dimension across stylized organizational types[2]). In an organization with high ARI and low modularity (A), subcommunity membership is stable and communication between subcommunities is frequent: these are classic matrix organizations where employees tend to work within one functional area but communicate frequently with others outside their area (e.g. Burns and Stalker, 1961). In an organization with high ARI and high modularity (B), subcommunity membership is stable and subcommunities are siloed: such an organization might be a functional hierarchy where individuals stay within their divisions and do not communicate across divisions (e.g. Sloan, 1963). An organization with low ARI and low modularity (C) might be post-bureaucratic (e.g. Kellogg et al., 2006) or a holocracy (e.g. Lee and Edmondson, 2017), with employees moving dynamically between and communicating across subcommunities. Finally, in an organization with low ARI and high modularity (D), employees move between subcommunities but those subcommunities are siloed: this might, for example, be a hospital emergency room (ER) where ER nurses and doctors work in intensely in temporary groups that change frequently depending on scheduling and shifts, but do not communicate with other groups while on-the-job (e.g. Valentine and Edmondson, 2014).

[2] While we use the terms "high" and "low" ARI and modularity for simplicity, in our data both ARI and modularity across organizations are quite high. In the context of Table 1, "high" and "low" should therefore be interpreted as relative rather than absolute terms. That is, an organization with "low" modularity might have lower-than-average modularity, rather than a low modularity score.

-- Insert Table 1 About Here –

While organizational scholars have identified consistently high modularity (0.60–0.75) in the *inter*-organizational collaboration networks of firms in several industries, including television (Clement et al. 2018), microelectronics (Tatarynowicz et al. 2016), computing (Gulati et al. 2012, Sytch and Tatarynowicz 2014) and biotechnology/pharmaceutical (Tatarynowicz et al. 2016), to our knowledge, research has not measured ARI or modularity in *intra*-organizational communication networks. In addition, studies have not explored temporal shifts in ARI and modularity (including those due to Covid-19). We attribute this gap to data limitations, as a comparative and dynamic analysis of network structure requires longitudinal data within and across many organizations. We leverage just such data to calculate changes in community structures following Covid-19.

**DATA**

We collected anonymized email data from approximately 100,000 organizations—approximately 450 billion email receipts[3]—over 29 months[4], from July 2018 through November 2020. The time period allows for both month-to-month (e.g. April 2019 to May 2019) and month-over-month (e.g. April 2019 to April 2020) comparisons that exploit Covid-19 as a natural experiment while accounting for seasonal variation (i.e. seasonality) in email patterns. For each organization $i$ and each month $t$, we constructed an undirected weighted edge $(u,v)$ with the weight $w_{i,t,(u,v)}$ being the number of messages observed between accounts $u$ and $v$[5]. To filter out companywide messages, we follow prior research (Kleinbaum et al. 2013) by not considering emails with more than four recipients.[6] We eliminate self-loops by ignoring edges from $u$ to $u$. This edge definition induced an undirected weighted graph from which we extracted the largest (weakly) connected component, denoted $G_{i,t} = (V_{i,t}, E_{i,t})$, where $V_{i,t}$ is the collection of accounts and edge $(u, v) \in E_{i,t}$ indicates that accounts $u$ and $v$ had at least one message between them. We limited our sample to organizations $i$ with $|V_{i,t}| > 2000$ for all $t$, yielding 4,361 organizations (ranging from 2,000 to 500,000 nodes), 126,469 organization-month networks, and approximately 362 billion email receipts.[7]

Using the same method, we analyzed data on multiple modes of communication between January 2019 and December 2020 from a single organization, Microsoft. After filtering out emails, instant messages, video calls, and scheduled calendar meetings with more than four recipients, we analyzed data on 524 million email receipts, 2 billion instant messages, 70 million video calls, and 39 million scheduled calendar meetings within the company over 24 months.

[3] We analyzed workplace trends that were anonymized by aggregating the data broadly. We neither observe nor use individual- or organization-level content, including information within an email or instant message. All personal and organization-identifying information, such as company name, were removed from the data before analysis.
[4] We focus on monthly periods as they allow us to balance richness and noise in communication. Shorter time periods, such as daily or weekly metrics, might be more granular but would be subject to more noise as communication patterns may differ across days of the week due to standing meetings, weekly workflow, etc. Longer time horizons such as yearly or quarterly periods would be subject to less noise, but would also compress information and sacrifice richness. While different time horizons may somewhat dampen or accentuate the results, the basic pattern of results should remain consistent.
[5] We use weighted edges because they allow for measurement of the *amount* of communication (rather than just the presence or absence of communication) between two nodes and give more signal to the graph. Using unweighted edges would increase noise and decrease our ability to meaningfully interpret analyses. To construct communities, it is critical to distinguish between employees who communicate multiple times per day on a regular basis and those who are only in periodic contact. As an example, using unweighted edges, it would appear as though employees are equally connected the CEO of the company (from whom they receive one email in a month) as to members of their team (with whom they email daily).
[6] We also analyzed the data by restricting emails to those with fewer than 2, 3, and 7 recipients; these specifications did not meaningfully affect our results.
[7] A cutoff of organizations with 2,000 or more email accounts was chosen to identify networks large enough for analysis, with relatively small standard deviations and distribution of modularity scores. Networks with fewer nodes are more sensitive to small changes in email patterns, making it harder to interpret shifts in community structure.

**ANALYSIS**

Analyzing the data was computationally intensive and required large-scale distributed compute infrastructure; it took more than 55,000 compute-hours for clustered machines. We defined modularity (Bickel and Chen 2009, Newman 2006, Newman and Girvan 2004) as $Q(G)$ = max 1 ∑ ($A$ − $dudv$ )$I${τ = τ }, where $A$ denotes the adjacency matrix, $d$ denotes τ$L$ $u$,$v$∈ $V$ $u$,$v$ $L$ $i$ $j$ $v$ the vertex degree, $L = \sum_v d_v$, and $\tau \in [K]^n$ encodes a network partition assigning $n$ vertices to $K$ communities. We used the Leiden algorithm (Traag et al. 2019) to find a network partition that approximately maximized the modularity function. For our sample of 4,361 organizations, we calculated the modularity scores of 126,469 organization-month networks using email receipts.

We also used email receipts for the entire sample to calculate ARI (Hubert and Arabie 1985). Given two networks on the same set of $n$ nodes at two different times, $G_t$ and $G_{t'}$, and letting $P_t$ and $P_{t'}$ be partitions of the two networks into communities, the Rand index is defined as $RI(G_t,G_{t'})=(a+b)/\binom{n}{2}$, where $a$ is the number of pairs of nodes that are in the same subset in both $P_t$ and $P_{t'}$ and $b$ is the number of pairs of nodes that are in different subsets in both partitions. ARI adjusts this measure for chance so that $ARI \approx 0$ indicates that nodes cluster together essentially by chance across the two networks, while $ARI \approx 1$ indicates that individual nodes' community memberships are stable across the two networks. We calculated ARI using the maximum modularity partitions.

To better illustrate the dynamic interplay between modularity and ARI, Figure 1 presents a simplified case: the network of a single organization with K = 2 blocks (communities), n = 12 vertices (nodes), and 6 vertices per block, observed at times G1 and G2. In stochastic block models (SBMs) (Holland et al. 1983) with both the number and size of the blocks held constant, an increase in Q together with low ARI implies (a) more siloed subcommunities and (b) significant instability in subcommunity membership. Figure 1 compares G1 and G2. The only difference between the two SBMs is in the block connectivity matrices B1 and B2, which are of the form [b11,b12; b21,b22]. We assume that within-block connectivity is b11 = b22 = 0.50 for both B1 and B2, but that between-block connectivity is b12 = b21 = 0.15 for B1, decreasing to 0.05 for B2. In this case, the modularity Q(G2) is larger than Q(G1): Q(G1) = 0.302 ± 0.065 versus Q(G2) = 0.404 ± 0.064. If we also assume that the block memberships are altered such that two of the 6 members of block 1 from G1 switch to block 2 in G2, replaced by two from G1's block 2 moving to G2's block 1, then ARI(G1,G2) = 0.02, as illustrated in Figure 1. The shaded communities are the Leiden-derived maximum modularity communities. The vertex colors denote block membership in G1; we can see that two vertices change communities. The modularity increase indicates sioloing of the internal community structure and the decrease in ARI indicates increased changes in community membership. We provide additional details, including the adjacency matrices, on the calculation of modularity and ARI for this example in the Appendix.

-- Insert Figure 1 About Here –

To further illustrate, Figure 2 presents a network map of email receipts at Microsoft in March 2020. The top image shows the entire organization ($n$ = 80,690) with modularity $Q_{Microsoft,MAR2020}$ = 0.82. Colors distinguish communities as discovered via the maximum modularity partition. We used formal organizational charts to isolate sub-organizations and present one with low modularity (left SubOrg 2, with $n$ = 10,243 and $Q$ = 0.79) and one with high modularity (right SubOrg 5, with $n$ = 29,958 and $Q$ = 0.85).

-- Insert Figure 2 About Here –

For the analysis of different communication modalities at Microsoft, we calculated monthly modularity scores and ARI for each modality (emails, instant messages, video calls, and scheduled calendar meetings).

**RESULTS**

We first examined changes in month-to-month modularity and ARI in 2019 and 2020 for organizations around the world using email receipts. Figures 3 and 4 provide a cross-sectional view of modularity across organizations and contextualize a shift in modularity within Microsoft. We find that modularity within organizations is relatively high: 50 percent of all organizations fall within the 0.64–0.77 range. Figure 3 provides a histogram summarizing the modularity $Q(G_{i,t})$ for all 126,469 organization-month networks, with interquartile ranges. Figure 3 shows email modularity within Microsoft at four times: February and March 2019 (red) and February and March 2020 (purple). Figure 4 is a heatmap for a kernel density estimate of network size $n$ versus modularity $Q$ for the 126,469 organization-month networks.

-- Insert Figure 3 About Here --

-- Insert Figure 4 About Here --

Figure 5 demonstrates monthly year-over-year modularity (mean ± 1 standard error) and ARI (mean ± 1 standard error) for 126,469 organization-month networks in 2019 and 2020. Figure 6 depicts the volume of emails sent and received by employees over time (where email volume indicates the average total number of emails an individual sent and received in a month); we see a sustained increase in March 2020, following the imposition of Covid-19 restrictions.[8] Figure 5 demonstrates a persistent increase in modularity and decrease in ARI[9] after March 2020.

-- Insert Figure 5 About Here --

-- Insert Figure 6 About Here --

Figure 7 presents an additional analysis of the March 2020 modularity increase apparent in Figure 5. The green histogram illustrates the year-over-year paired difference in modularity between January 2019 and January 2020, centered near 0 (indicating no statistically significant difference in modularity). The red histogram illustrates the year-over-year difference between April 2019 and April 2020, centered greater than 0 (indicating increased modularity in 2020; two-sided Wilcoxon $p$-value ≈ 0), showing a post-Covid-19 effect.

[8] As in prior studies (DeFilippis et al. 2022), we observe that the volume of emails decreased following the steep spike in March 2020; we build on these studies by showing that this decrease was not sustained and that the volume of emails remained higher throughout 2020 than it had been in 2019.

[9] We only calculate ARI for nodes that existed in two periods. This helps avoid measuring instability driven by employees leaving an organization entirely, rather than nodes moving between two subcommunities within the organization. Moreover, we do not observe significant changes in the number of nodes (employees) within organizations, further supporting the idea that the observed ARI shifts were not due to employees leaving organizations.

-- Insert Figure 7 About Here –

**Additional Analyses**

**Alternative Communication Modalities:** While our findings indicate shifts in the modularity and ARI of email communication, there are two alternative explanations for our results. The first is substitution: perhaps global workplace email communication appeared to become more siloed or membership appeared to become less stable because in-person communication networks had been more siloed and less stable than email networks in 2019 than in 2020. If so, perhaps the modularity and ARI of overall communication did not change following Covid-19, but the shift to remote work simply allowed us to better observe an organization's rates of siloing and stability. The second explanation for our results is true dynamic siloing: perhaps the communities observed in intra-organizational communication networks fundamentally shifted due to the rise of remote work.

To evaluate these explanations, we examined shifts in volume, modularity, and ARI across four modes of communication (emails, instant messages, video calls, and calendaring software) within Microsoft following the imposition of a companywide work-from-home order on March 4, 2020. First, we examined shifts in volume across these modalities to understand which of them may have substituted for in-person communication. Figure 8 illustrates the change in volume within Microsoft for each modality: in 2020, the volume of emails and calendar meetings increased initially but quickly returned to 2019 levels, while the volume of instant messages and video calls increased significantly over the long term. Our findings are consistent with Yang et al. (2022), who noted a sustained increase in instant messages and video calls and short-lived increases—followed by readjustment—of email and meeting volume. This suggests that, following Covid-19, instant messages and video calls at Microsoft substituted for in-person interactions.

-- Insert Figure 8 About Here –

Next, we calculated whether and how modularity and ARI shifted within each communication modality. If shifts are driven by substitution rather than a change in communication patterns, we would expect modularity and ARI to stay the same for emails and calendar meetings (because these do not appear to have substituted for in-person interactions) and increase for instant messages and video calls. We find instead that modularity and ARI increased for all modalities, even those where substitution did not appear to be occurring. Using email receipts, we find that $\Delta Q(GMicrosoft,Feb2019, GMicrosoft,Mar2019) = Q(GMicrosoft,Mar2019) - Q(GMicrosoft,Feb\ 2019) = -0.001$, while $\Delta Q(GMicrosoft,Feb2020, GMicrosoft,Mar2020) = 0.013$. Figure 9 plots modularity (left panel) and ARI (right panel) for Microsoft email networks from January 2019 through December 2020 and shows the modularity increase and ARI decrease that followed the work-from-home order.

-- Insert Figure 9 About Here --

Figure 10 plots modularity (left panel) and ARI (right panel) within Microsoft for the other modalities. The results are consistent across modalities. Using instant messages, we find that $\Delta Q(GMicrosoft,Feb2019, GMicrosoft,Mar2019) = Q(GMicrosoft,Mar2019) - Q(GMicrosoft,Feb\ 2019) = -0.024$, while $\Delta Q(GMicrosoft,Feb2020, GMicrosoft,Mar2020) = 0.025$. Using video calls, we find that $\Delta Q(GMicrosoft,Feb2019, GMicrosoft,Mar2019) = Q(GMicrosoft,Mar2019) - Q(GMicrosoft,Feb\ 2019) = 0.000$, while $\Delta Q(GMicrosoft,Feb2020, GMicrosoft,Mar2020) = 0.036$. For calendar meetings, we find that $\Delta Q(GMicrosoft,Feb2019, GMicrosoft,Mar2019) = Q(GMicrosoft,Mar2019) - Q(GMicrosoft,Feb\ 2019) = -0.004$, while $\Delta Q(GMicrosoft,Feb2020, GMicrosoft,Mar2020) = 0.012$.

To assess the significance of the modularity change in email networks from February to March 2020, we consider network bootstrapping, which allows us to quantify the uncertainty in inferences about network statistics (Green and Shalizi 2022). For February 2020, the observed modularity $Q = 0.807$ and the bootstrap yields $Q = 0.804 \pm 0.0037$; for March 2020, the observed modularity $Q = 0.820$ and the bootstrap yields $Q = 0.818 \pm 0.0035$. This provides evidence that communication patterns within Microsoft fundamentally changed as employees shifted to working from home.

-- Insert Figure 10 About Here --

**Partitioning Algorithm vs. Formal Organization Structure.** There is also a possibility that our choice of partitioning algorithm for modularity inherently captures modularity where there is none because it is based on maximum partitioning. We examine whether our choice of algorithm is driving the results in several ways. One alternative to using algorithmic maximization techniques to identify (i.e. infer) communities is using the formal (i.e. imposed) organizational structure to partition the network[10]. We explored this possibility by using formal organization structure to analyze shifts in email modularity for seven sub-organizations within Microsoft. These subunits had different initial modularity scores, likely driven by their distinct strategies, approaches to work, and degree of coupling with the broader organization. We found that our results held in all but one sub-organization. That one, which dropped in modularity significantly between February and March 2020, included members of Microsoft's strategy and operations organization tasked with manning the organization's control center through the crisis. Members nimbly adapted their networks by paring down connections within their own working group and by keeping and forming connections across groups—likely those most acutely relevant. This indicates that shifts in email communication, while broadly similar between organizations, likely differed between communities within the same organization based on their function, strategy, and degree of coupling to the rest of the organization.

To further explore the validity of using Leiden, we obtained data on formal organization structures within all of Microsoft across the 24-month period to impose—rather than infer—community structure. As a large dynamic organization, Microsoft has significant inflows and outflows of employees and frequently reorganizes workgroups and functional areas; we therefore used the monthly organizational charts to capture these dynamics. We used the formally defined organizational hierarchy to define a new partitioning scheme and observed the changes in modularity and ARI pre- and post-Covid and compared these results to those from our original analyses. Using the imposed organizational structure as the partitioning scheme, we generated Figure 11. We observed the same trends as in Figure 9, indicating that the findings using formal organization structure are consistent with the findings based on inferred (Leiden) structure.

[10] We note that using a partitioning algorithm like Leiden avoids potential biases and arbitrary decisions related to partitioning, captures actual communication flows (among functions, between collaborators, at different levels of hierarchy, etc.), that may be missed with imposed network partitions, adjusts dynamically to shifting communication patterns, and can alleviate privacy concerns that attend the use of sensitive communication data. There is also some evidence that imposed partitions, such as formal organizational structures, may not accurately capture work functions or who communicates or collaborates with whom (e.g., Gulati and Puranam 2009, Kleinbaum and Stuart 2014).

-- Insert Figure 11 About Here --

Next, we compared the formal organizational-structure–based partitioning to the inferred organizational structure using the "Freedom" metric which "measures the expressed freedom of communities to form across organizational boundaries" (Edge et al. 2021). To calculate alignment between a community and a formal organizational structure, we constructed a minimum spanning tree (MST) that connects all members of a community across the formal organizational structure. If the community and the formal organizational structure are perfectly aligned, then this MST would consist entirely of community members (there would be no division of reporting lines in this case). In the formal structure, all peers of all MST nodes would also be members of the same community, with the exception of the root note of the MST. In this case, peers are not members of another community, which enforces a nondivision of peer groups except for team leaders. Using these criteria, we created a continuous alignment scale and penalized the division of reporting lines and peer groups and calculate alignment – that is, Freedom – based on it. Formally, for a community *c*:

$$Freedom_c = 1 - \frac{|c|}{|MST_c \cup \textit{peers of non-root nodes in } MST_c|}$$

Figure 12 represents the calculation of Freedom for two organizations with 16 vertices each. It illustrates differing degrees of alignment between the formal organization structure and actual communication structure. In Example 1, there is perfect alignment between formal structure and actual communication (that is, individuals only communicate with others in their community) and Freedom equals 0. In Example 2, there is less alignment; individuals communicate across communities, causing Freedom to increase to 0.5

-- Insert Figure 12 About Here --

Applying this method to the Leiden communities and to Microsoft's formal organizational hierarchy, we generated Figure 13. In 2019, Freedom ranged between 0.48 and 0.58, indicating that (a) for most months, a substantial amount of email was between individuals in different organizational subcommunities or (b) individuals in formal subcommunities (e.g. workgroups) did not communicate via email. Beginning in April 2020, the first full month of work-from-home orders at Microsoft, Freedom went down as adjustments to Covid were put in place, hitting a minimum in June 2020.[11] The decreases in Freedom observed in 2002 relative to 2021 indicates that employees became more siloed in their communications and communicated more within their formal organizational hierarchies and less with other subcommunities. We observe that Freedom was lower post-Covid, suggesting that individuals were much more likely to communicate via email *within* their workgroup post-Covid. This supports the idea that communities within organizations were indeed more siloed post-Covid.

-- Insert Figure 13 About Here --

We also ran this analysis on a per-community level (rather than for all of Microsoft) to examine the distribution of Freedom (Edge et al. 2021). In Figure 14, each derived community is represented by a dot, the size of which represents the size of the community. We noted considerable variation in the alignment between the imposed and inferred communities. This disconnect suggests that imposed partitions may not adequately capture all communication within an organization, and thus supports our use of Leiden.

-- Insert Figure 14 About Here --

[11] December 2020 is excluded from our calculations because we could not access the human resource data needed to calculate Freedom for that month.

Overall, the additional analyses on multiple communication modalities, alternative partitioning schemes, and comparison to formal organization charts each produced results consistent with the main analyses.

## GENERATIVE MODEL

The study of intra-organizational networks is often prohibited by issues related to access, computational intensity, privacy, and legal considerations that prohibit obtaining raw communication data for analysis. To facilitate future research, we created a generative model designed for our intra-organizational communication networks and have negotiated to make the code for this model, along with the anonymized data on Microsoft, available to other scholars.

Our model is a Barabási-Albert augmented hierarchical stochastic block model (BA-HSBM), a generalization of the Barabási-Albert generative model (Barabási and Albert 1999). Figure 15 presents a visual comparison of this new model with its simpler competitors. We provide a fitted generative model for all 126,469 organization-month networks, allowing for future research and providing baselines for comparisons against observed activity.

-- Insert Figure 15 About Here --

To generate this model, we used root-level Leiden community structures to create an a-posteriori stochastic block model that retained the population statistics for vertices and edges from the real network being fit. To make Barabási-Albert fit well in the context of an SBM, we modified the algorithm: within each block of the SBM, we considered a specific budget of vertices and edges, obtained from the observed network being fit. We configured the Barabási-Albert algorithm to create a number of edges for each vertex equal to the intra-block average degree centrality. Then, using either Erdös-Rényi (Figure 15, Panel (a)) or Barabási-Albert (Figure 15, Panel (b)) to create intra-block connections, we observed major differences between the resulting networks and the real network being fit (Figure 15, Panel (d)). The interblock connections are made at a rate determined by the real network, but they are made between random vertices across pairs of communities. We observed that the power-law distribution of the degree centrality is much closer to the real network's distribution when using the Barabási-Albert generator and that the network paths generated using Barabási-Albert are longer than those from Erdös-Rényi. These longer paths produce less regularity in the structures and allow for bleed-over between communities, as highly eccentric nodes connected to multiple communities will be pulled between those communities.

Using these observations, we extended our model to use hierarchical Leiden communities obtained by running Leiden recursively on the real network until we attained leaf communities with no more than *nmax* vertices. (We use *nmax* = 250.) Using these leaf communities, we applied the Barabási-Albert algorithm again for the leaf intrablock connections, then proceeded with interblock connections between leaf clusters. This localized the connections between communities to small groups of nodes, dramatically fracturing the network structure (Figure 15, Panel (c)) and corresponding to the structure observed in the real network being fit; as in the real network, the generative model produced many new and small communities. Applying Leiden to data generated from our generative model, we found these groups of communities captured in the same partition when maximizing root-level modularity, indicating that the more complex and realistic structure generated by BA-HSBM has modularity characteristics similar to those of the real network being fit. This generative model significantly reduces the computational complexity of analyzing the data in the future.

## DISCUSSION

Our analyses indicate that the modularity of intra-organizational communication networks increased from 2019 to 2020, while ARI decreased. These results were replicated for organizations around the world and across multiple communication modalities for one organization, Microsoft. As employees shifted to remote work due to Covid-19, organizational networks around the world became more siloed and the membership within communities within these networks became less stable. The widespread shifts in these measures—that is, dynamic silos—suggest that remote work fundamentally reshaped intra-organizational community structures.

The term *dynamic silos* captures the counter-intuitive idea that subcommunities (whether teams, workgroups, and functional areas) can become both *less* stable in their membership and *more* siloed in their communication patterns over time. Our results indicate that, following the rise of Covid-related work-from-home-orders, employees began to switch more dynamically between teams, workgroups, or functional areas. Scholars have argued that membership instability can arise in subcommunities or teams "that face complexity, dynamism, or novelty because they must accommodate time-sensitive work, competing demands, or emerging needs that make it difficult to anticipate the resources, skills, and roles that will be required" (Kerrissey et al. 2020: 4). The Covid-19 crisis created precisely these sorts of conditions, and demanded greater flexibility from both organizations (that had to adapt quickly to dramatic changes in their environment including demand shocks) and their employees (who had to shift their work tasks and hours because of household responsibilities including disruptions to schooling, childcare, etc.) (Alekseev et al. 2022). Greater flexibility may have resulted in decreased stability, as employees moved between subcommunities more fluidly to accommodate environmental, organizational, and personal changes.

At the same time, we found that, once *in* a team, workgroup, or functional area, employees increasingly limited their communication to other members of that subcommunity. This finding builds on prior studies showing that, following a shift to remote work, employees communicated less frequently with a smaller number of weak ties (Yang et al. 2022). This might be because their communication became more essentialist or focused on work tasks, or because serendipitous, in-person interactions with those outside one's community were not replaced by digital communication. To our knowledge, ours is the first study of whether and how these individual-level effects are aggregated into shifts in an organization's community structures. In our study, this essentialism may have resulted in increased siloing.

While we do not test it directly, an interesting implication of our findings is that, following Covid-19, membership in a subcommunity might predict communication patterns more strongly than the strength of past ties (because modularity increased even as employees moved more dynamically between subcommunities). Indeed, there is an assumption (often implicit) in organizational network research that ties are sticky and persist over time. But our results show that ties may not always remain active as employees move between communities. This is important for organizational leaders as they think about design and the future of remote work. For example, many organizations deliberately move people around different functional and product areas to promote learning and cross-pollination of ideas. But our results imply that in remote work settings decreased stability may not increase communication between different functional areas. The ability to disentangle modularity and stability as distinct dimensions of community structure, and to measure shifts in each, is a significant contribution of our study relative to prior research focused not on community structures but on individual-level outcomes (e.g. Gibbs et al. 2021; Yang et al. 2022). As organizational scholars increasingly gain access to large-scale communication data, accounting for the distinctiveness of communication patterns in remote work vs. traditional in-person work needs to be accounted for.

Our findings have additional implications for both theory and practice. Pre-Covid studies of remote work were largely optimistic about its potential, finding that work-from-home arrangements provided more flexibility, less distraction, and greater employee satisfaction and performance (e.g., Bloom et al. 2014, Gajendran and Harrison 2007). But, as noted, early post-Covid studies have highlighted the individual-level negative effects of remote work: longer workdays (e.g., DeFilippis et al. 2022), lower productivity (e.g., Gibbs et al. 2021), less contact with weak ties and fewer new connections (Yang et al. 2022), and fewer opportunities for coaching and mentoring (Gibbs et al. 2021). Recent research (Choudhury et al. 2021) has suggested that hybrid work arrangements consisting of both in-office and at-home work might overcome these challenges. Our findings highlight community-level changes that followed Covid-19 with implications for research on organizational performance and innovation and on intra-organizational networks.

Future research can explore both the benefits and the tradeoffs of dynamic silos. As Figures 1 and 2 illustrate, even small shifts in modularity and ARI produce intra-organizational communication networks that operate in dissimilar ways. For example, increased modularity might improve productivity and efficiency. More siloed networks might promote the rapid sharing of information and tacit knowledge (Granovetter 1985). Research has suggested that these benefits can occur even with membership instability, provided that members have clear roles within their groups (Valentine and Edmondson 2015). Thus, depending on the work practices and other features of collaborative groups, dynamic siloing may improve efficiency (Choudhury et al. 2021) and productivity (Dahlander and McFarland 2013). Future research can examine whether dynamic siloing is indeed associated with changes in firm or unit performance and whether or how these relationships are moderated by different work practices.

There is also the possibility that dynamic siloing may reduce innovation in some organizations. Innovation often arises from novel combinations of distantly held knowledge (Burt 2004, Hargadon and Sutton 1997, Kogut and Zander 1992, Schumpeter and Opie 1934). Interdisciplinary or cross-department collaborations provide access to new ties and information that can provoke innovative ideas (Rawlings et al. 2015, Soda et al. 2021). It is tempting to believe that shifting membership across workgroups – for example, through employee rotations between teams – can strengthen ties between subcommunities, increasing recombination and perhaps innovation. But our findings indicate that membership instability can – somewhat paradoxically – be accompanied by greater siloing. This siloing could reduce access to new knowledge even in the face of instability (Gulati et al. 2012, Tortoriello et al. 2012, Uzzi 1997). Future research should examine the impact of concurrent shifts in modularity and ARI on innovation (measured through patents, publications, new products, and so on).

Finally, dynamic siloing in large organizations might be associated with a specific kind of innovation; namely, competence-destroying technologies (Abernathy and Clark 1985, Tushman and Anderson 1986) that render existing organizational capabilities obsolete. Such innovations are typically the work of new firms (Tripsas 1997, Zuzul and Tripsas 2020) or small teams (Wu et al. 2019). In large or incumbent organizations, competence-destroying, architectural, or disruptive innovation is best developed by groups that are loosely coupled with the rest of the organization. As increased modularity might foster just such cultural separation and autonomy, dynamic siloing might promote innovation in large established organizations (Benner and Tushman 2003, Christensen 1997, Henderson and Clark 1990). Future research could examine the relationship between modularity and the *kind* of innovation produced by large firms.

This study is also one of the first to infer community structures within organizations and to compare the structure of intra-organizational communication networks over time. To be studied, communities—whether teams or subgroups—can be either exogenously defined through criteria like formal roles or inferred through observation of patterns of interaction (Katz et al. 2004). Scholars have demonstrated the importance of inferred community structures in understanding, for instance, how epidemics spread (Kleczkowski and Grenfell 1999) or how to best allocate tasks to parallel computing processors (Fortunato 2010). We found that, within Microsoft, inferred community structures were better than formal organizational charts as predictors of actual communication. We hope that our study—including our generative model—catalyzes research further inferring community structures within intra-organizational networks. We see significant potential in examining how these structures are formed and whether and how they change in response to environmental or strategic shifts. We also see the potential for research examining whether the degree of coupling between an organization's formal and inferred communities has implications for innovation and performance.

Although not the primary focus of our analysis, our findings also highlight the need for research on the use of digital technologies in organizations and on geographic differences in intra-organizational communication structures. First, we found that, following the imposition of Microsoft's work-from-home order, video calls and instant messages appeared to substitute for in-person communication. Media richness theory has posited that low-richness media like instant messaging are less effective than high-richness media like in-person conversations for managing complex interactions (Daft and Lengel 1986, Daft and Wiginton 1979, Lengel and Daft 1984), particularly those that involve emotions (Byron 2008, Maruping and Agarwal 2004). Some evidence suggests that, as a result, virtual or distributed teams—even those using videoconferencing—generate fewer creative ideas (Brucks and Levav 2022) and make fewer breakthrough scientific discoveries (Lin, Frey, and Wu 2022) than co-located teams. Future research can replicate our results across organizations and examine the implications of the substitution of in-person interactions with low-richness media.

Finally, our analysis highlights the need to examine the drivers and implications of geographic differences in baseline modularity scores. Although we do not present these results (because we did not examine random or theoretical samples of organizations across geographies), we noted suggestive differences between the modularity scores of organizations in different countries. An interesting extension of our work would be to formally consider why and how modularity might differ across regions. We also noted geographic differences in the magnitude of post-Covid-19 modularity shifts. If these shifts are associated with changes in organizational performance and innovation, they may have implications for national competitiveness and resilience and therefore merit continued focus as organizational communities evolve after the pandemic.

Practically, our research indicates that executives of organizations embracing remote work need to understand how such a change can affect the structure of intra-organizational communication networks. Executives who made formal organizational changes in response to Covid-19 must consider whether and how those changes need to be adapted to support long-term remote work. Ultimately, by understanding that shifts to remote work can affect not just how employees communicate, but also how intra-organizational communities are structured and connected, executives can begin to attend to the implications of dynamic silos.

An advantage of large-scale analysis of digital communication across thousands of organizations and more than a billion individuals is the ability to uncover high-level shifts in community structures. We cannot, however, observe or measure how these shifts affect organizational outcomes, nor can we observe in-person communications. These limitations of our study offer opportunities for future research. First, as with any empirical study, we had to identify the appropriate time periods to calculate and compare modularity and ARI. In using monthly calculations, we attempted to balance capturing seasonal effects that vary between longer periods with introducing noise by measuring shorter periods. Ultimately, our choice of monthly periods allows for year-over-year comparisons that likely capture seasonality while also capturing work practices that may vary between days or weeks in a month. Choosing longer or shorter time horizons could somewhat dampen or accentuate the results, but the basic pattern of results should be consistent.  Given that this is the first study to calculate modularity and ARI for intra-organizational communication, these results serve as important baseline for future comparisons. Future studies might examine how using longer or shorter time horizons impact observed modularity and ARI. They may also highlight variation in the appropriate time period for different industries, regions and communication modalities. Future studies on a smaller subset of organizations or even a single organization across less time can explore these and related questions as understanding modalities in remote work will likely become increasingly important as it continues to be experimented with by organizations.

The purpose of our exploratory study was to uncover broad patterns in the impact of Covid-19 on community structures within organizations. We hope future studies will test and refine our results; more generally, we hope our research will stimulate studies connecting community structures with organizational outcomes.

**Data Availability**

An anonymized version of the email data on Microsoft that support this study will be retained indefinitely for scientific and academic purposes. The data are available from the authors upon reasonable request and with the permission of Microsoft.

**Code Availability**

The code used to produce the results shown on Microsoft and the code used to create the generative models and the fitted generative models for all 126,469 organization-month networks is available upon reasonable request and with the permission of Microsoft.

| | Modularity: Low | Modularity: High |
|---|---|---|
| ARI: High | **A**<br>Subcommunity stability is high; subcommunities are connected | **B**<br>Subcommunity stability is high; subcommunities are siloed |
| | e.g. Matrix organization | e.g. Functional hierarchy |
| ARI: Low | **C**<br>Subcommunity stability is low; subcommunities are connected | **D**<br>Subcommunity stability is low; subcommunities are siloed |
| | e.g. Holocracy | e.g. Emergency room (ER) teams |

**Table 1: ARI and Modularity as Distinct Dimensions of Community Structure**

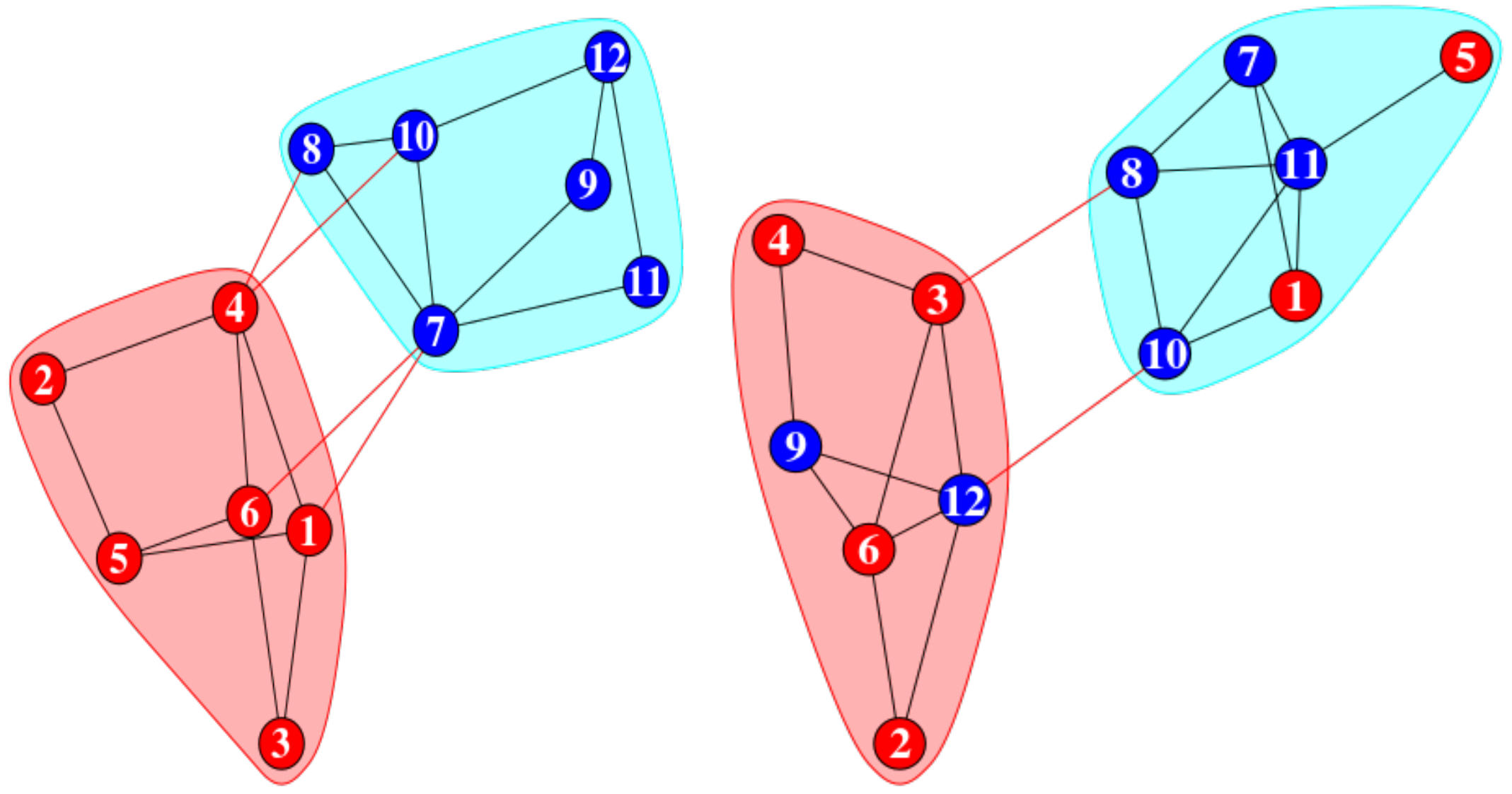

**Figure 1: A Simplified Case of Modularity and ARI Change in an Illustrative Network**

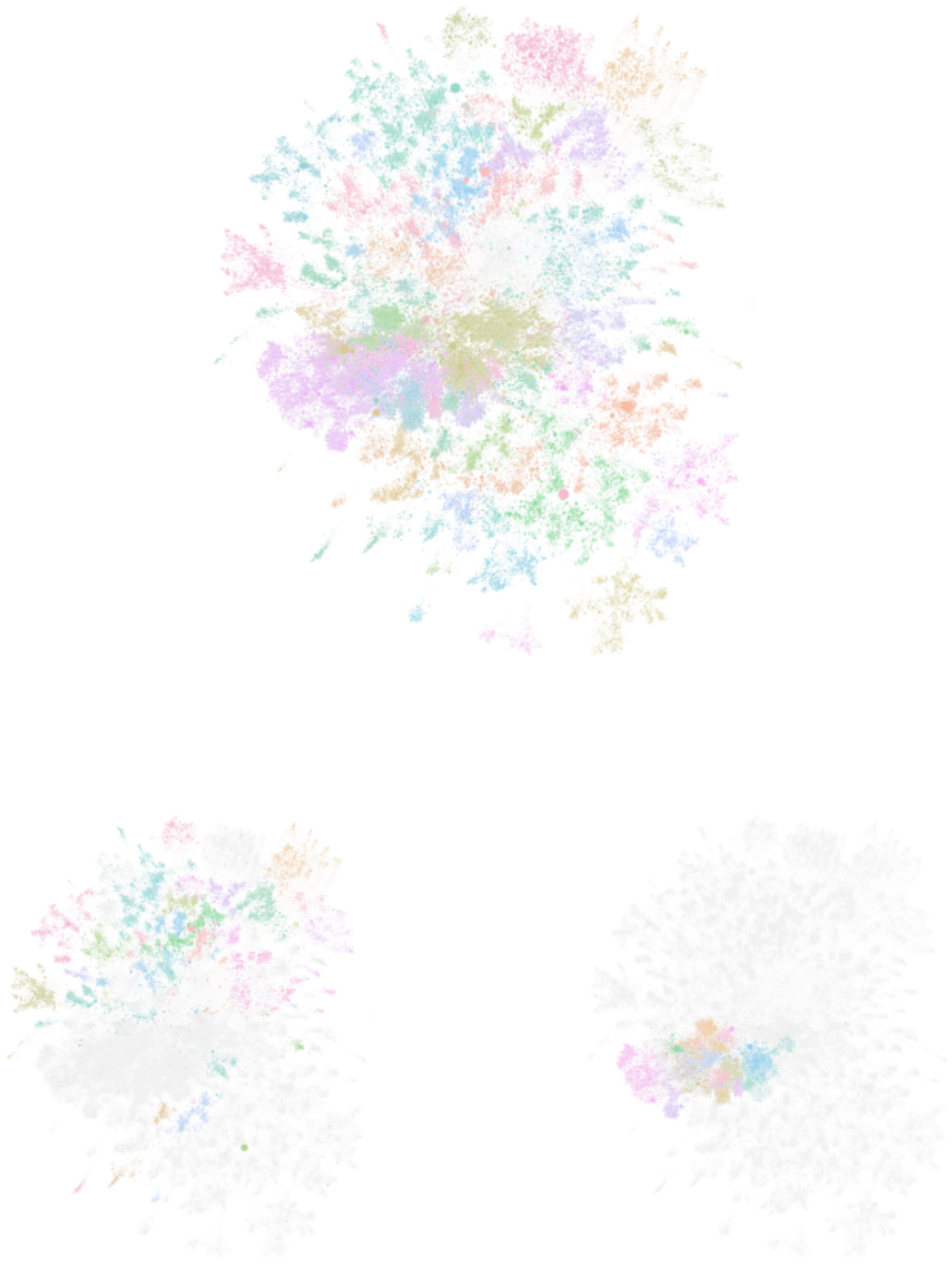

**Figure 2: Network Map of Email Receipts within Microsoft, March 2020.** Colors indicate communities discovered via the maximum modularity partition. Top image shows the entire organization (modularity = .82). Lower left image shows a low-modularity sub-organization (modularity = .79), identified via a formal organizational chart; lower right image shows a high-modularity sub-organization (modularity = .85), identified via a formal organizational chart.

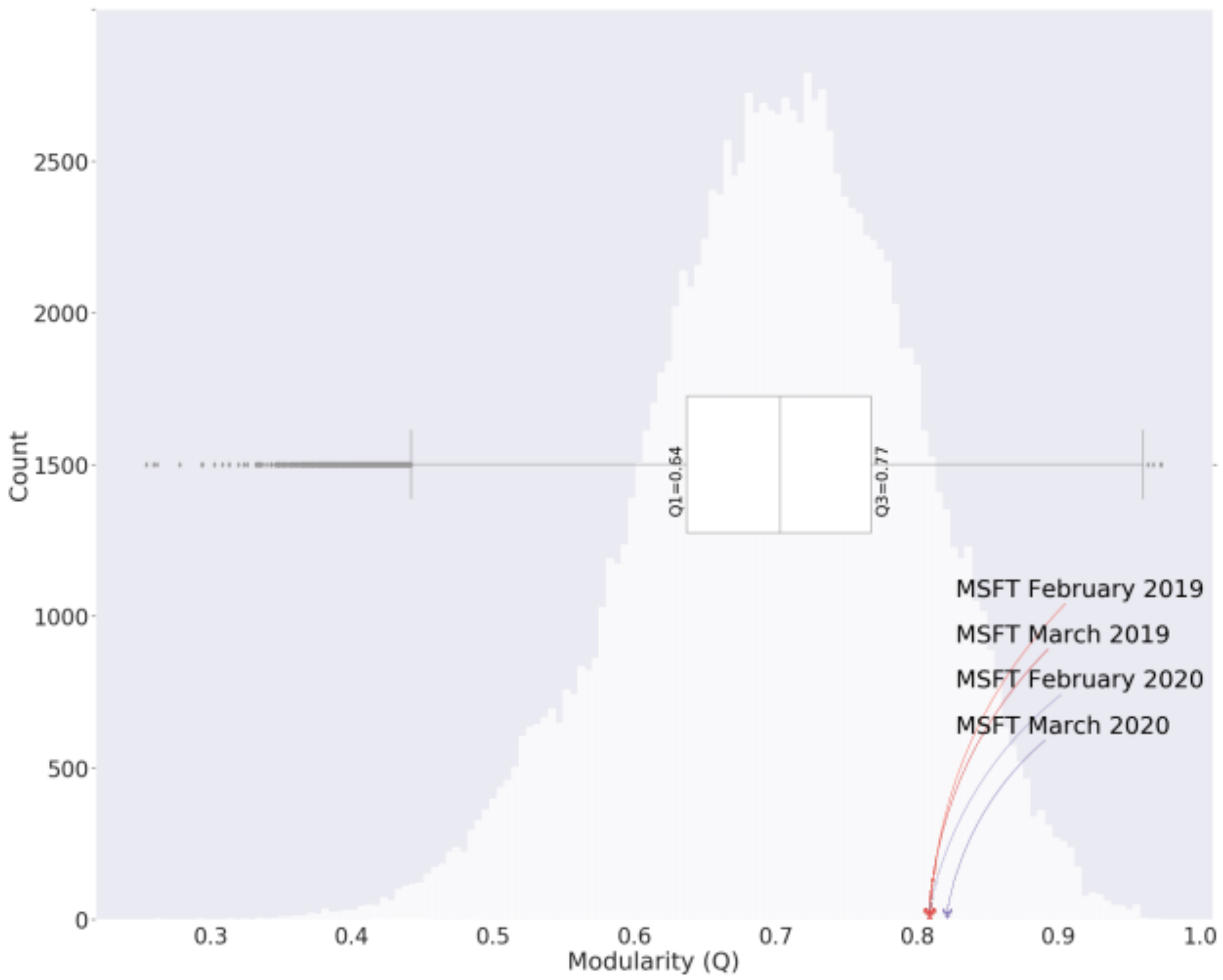

**Figure 3: Modularity Histogram and Boxplot.** Illustrating modularity for 126,469 organization-month networks and monthly differences within Microsoft.

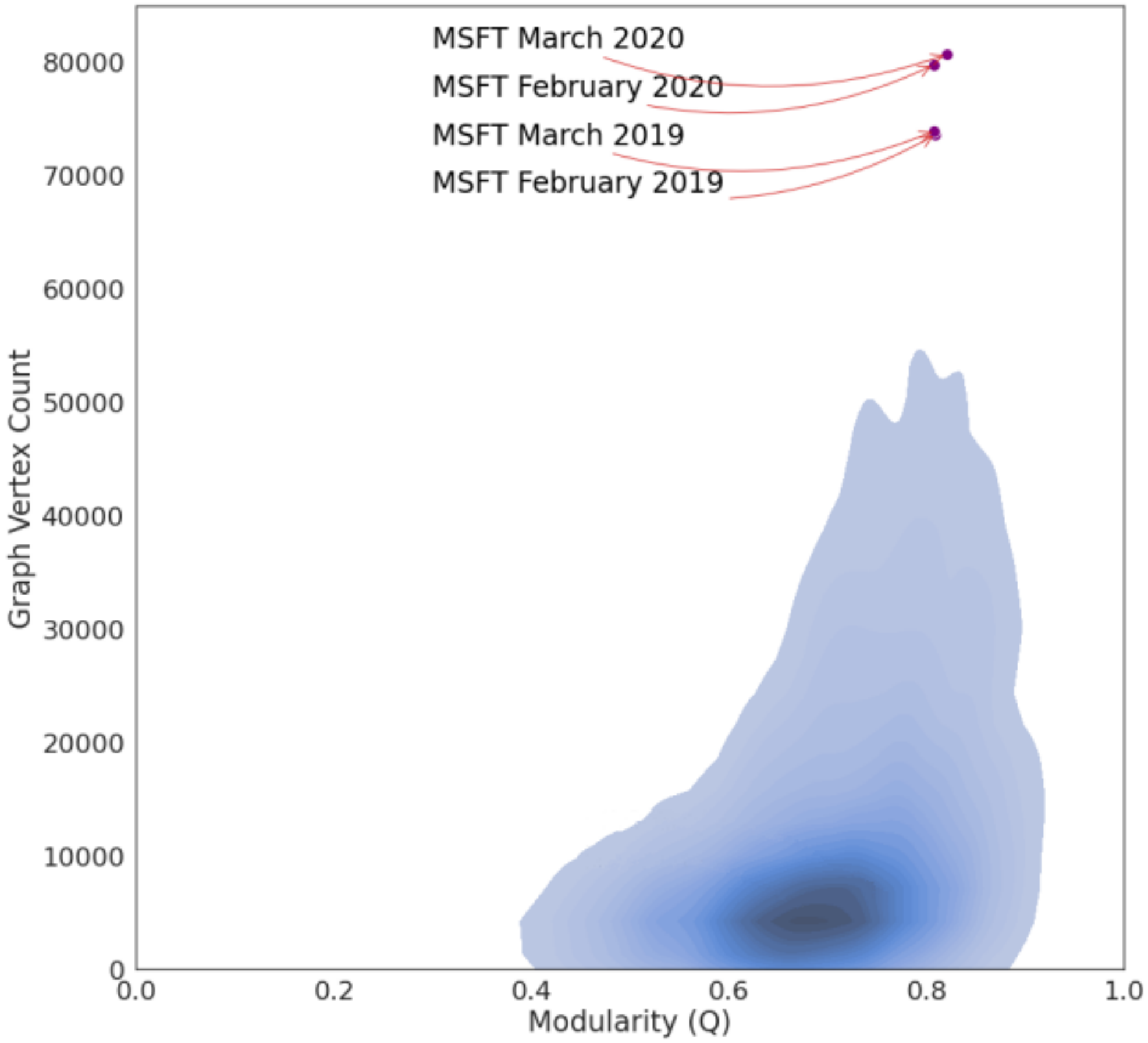

**Figure 4: Heatmap for a Kernel Density Estimate of Network Size vs. Modularity**. Illustrating monthly differences within Microsoft.

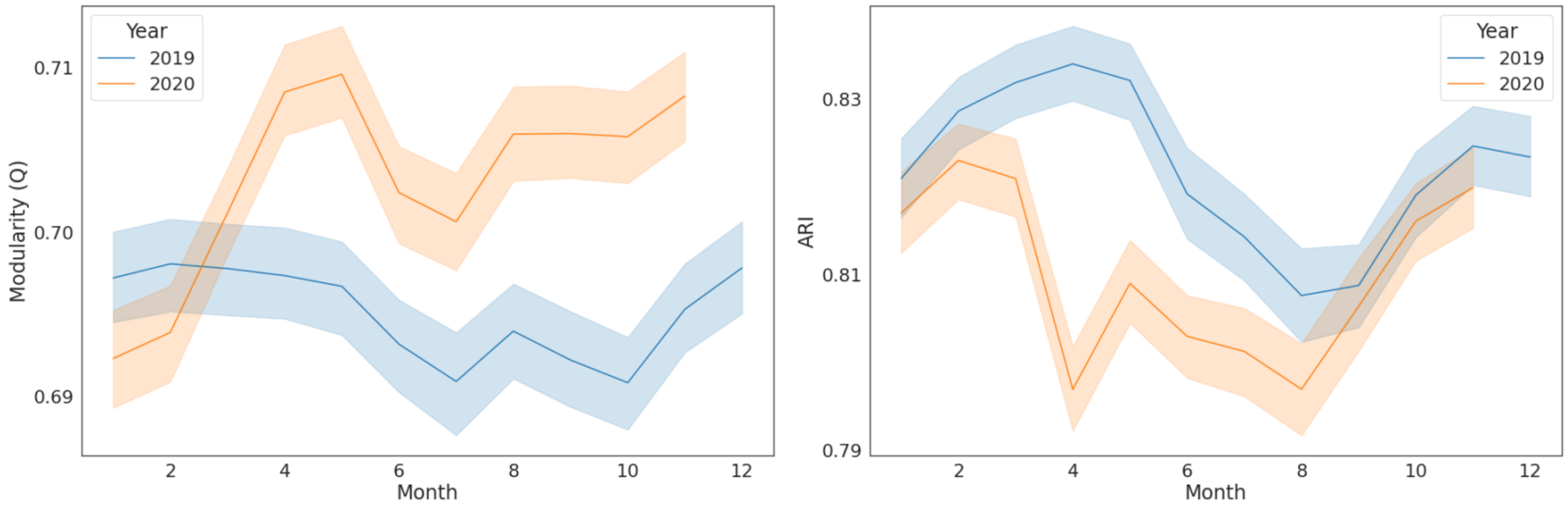

**Figure 5: Monthly Year-over-Year Global Modularity and ARI, 2019 vs. 2020.** Illustrating mean ± 1 standard error.

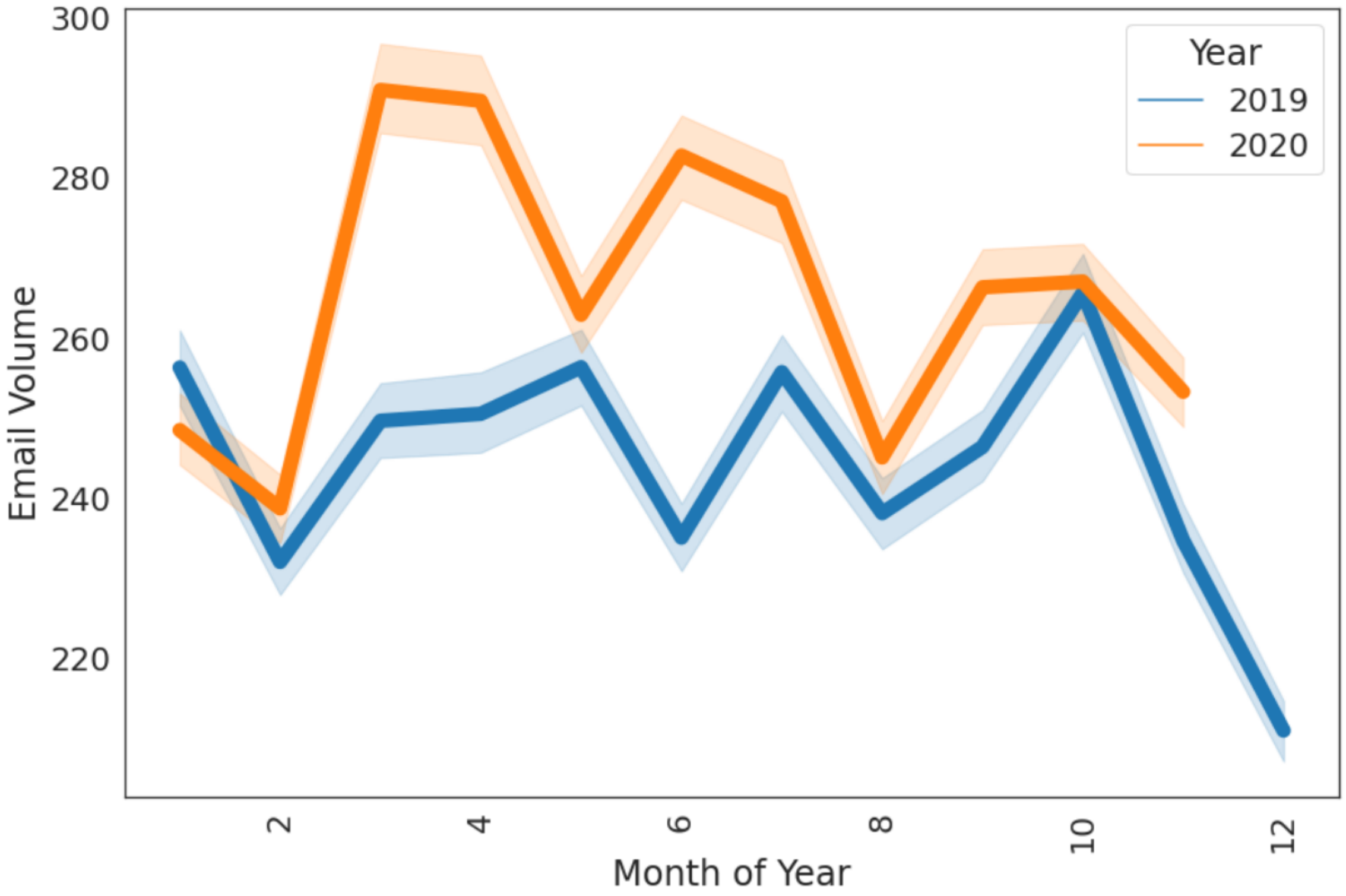

**Figure 6: Monthly Year-over-Year Global Email Volume, 2019 vs. 2020.** Illustrating mean ± 1 standard error.

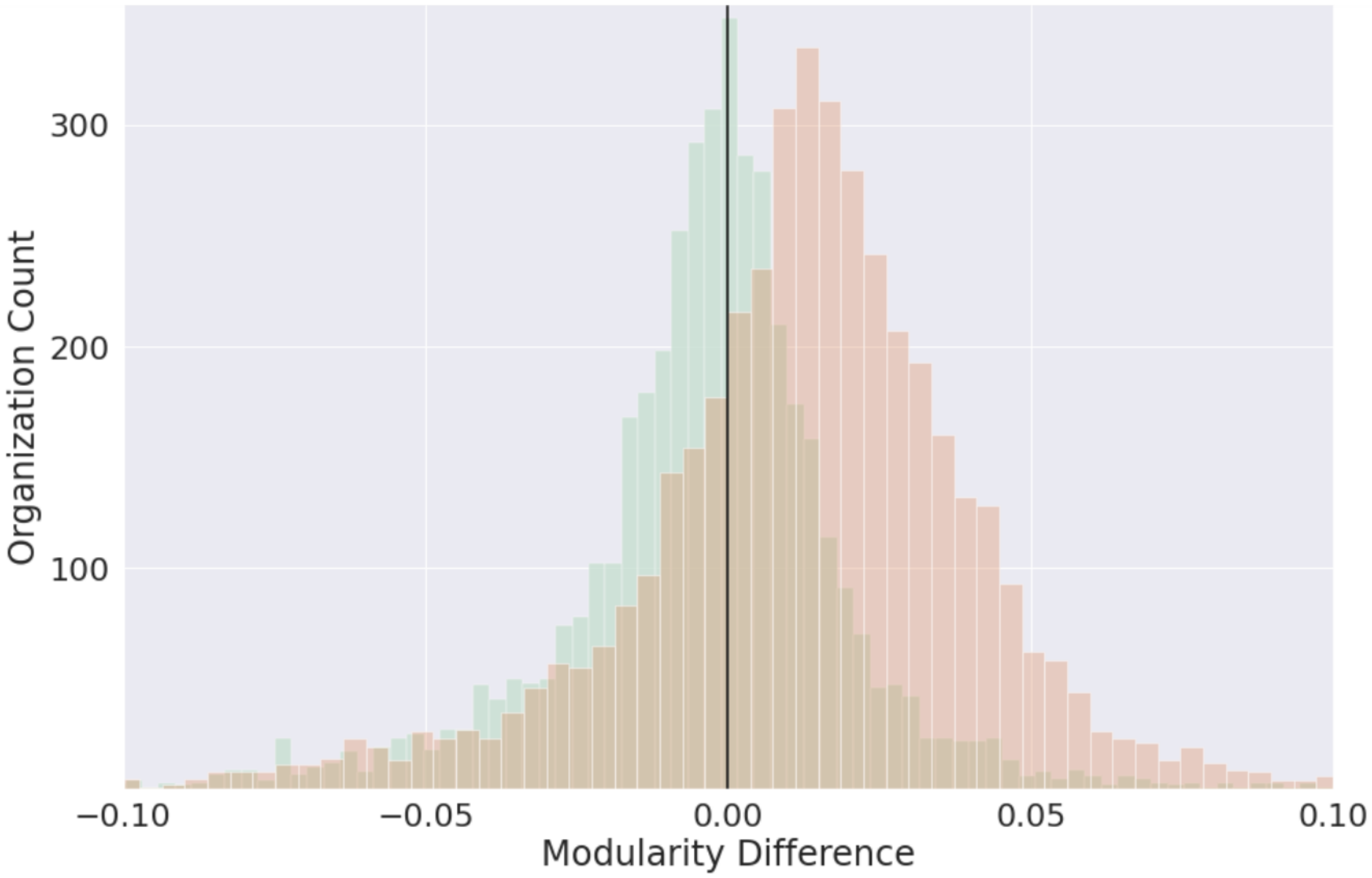

**Figure 7: Monthly Year-over-Year Global Modularity Paired Difference Histograms.** Green illustrates the difference in January 2019 vs. January 2020. Red illustrates the difference in April 2019 vs. April 2020.

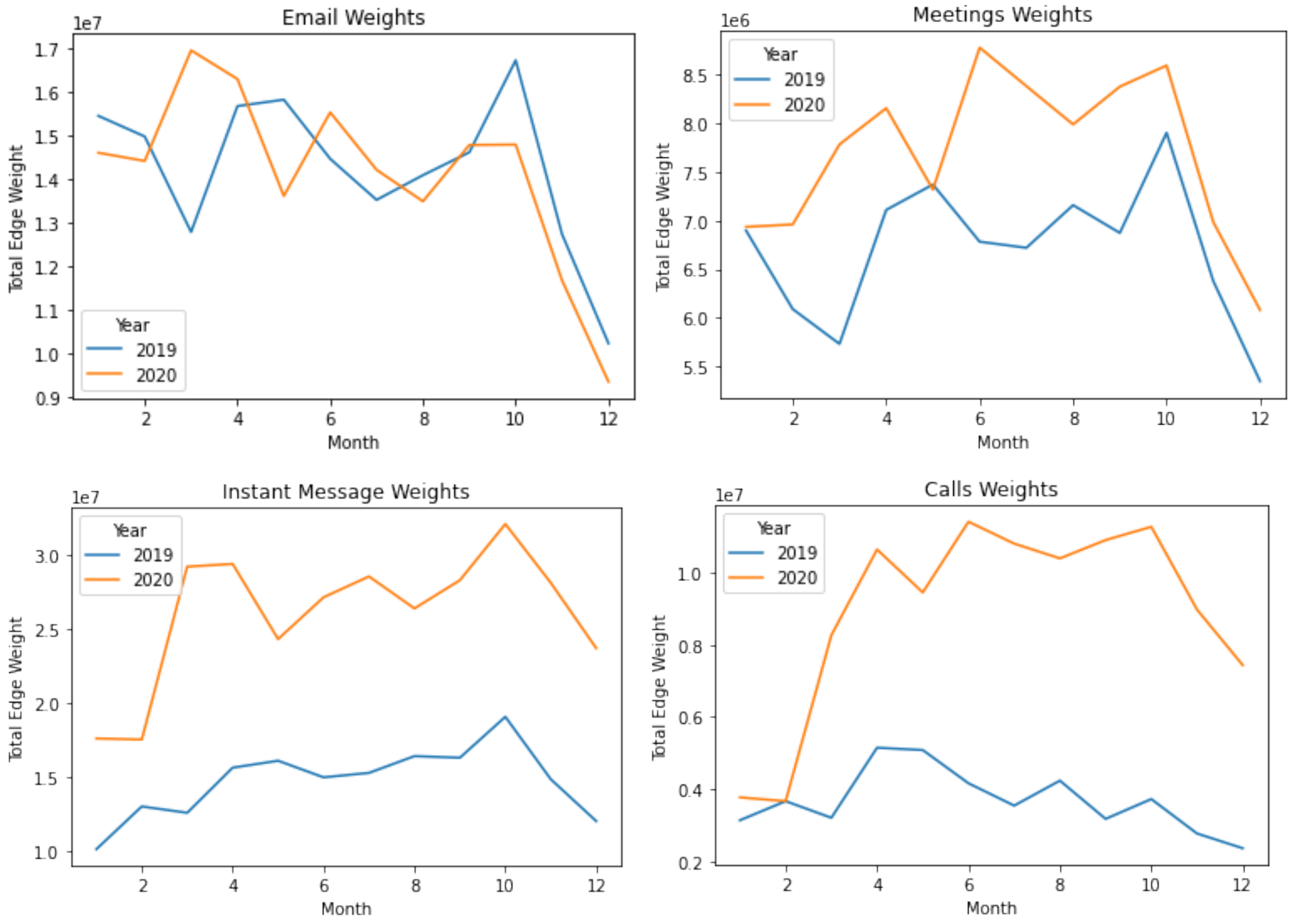

**Figure 8: Monthly Year-over-Year Microsoft Email, Meeting, Instant Message, and Video Call Volume, 2019 vs. 2020**

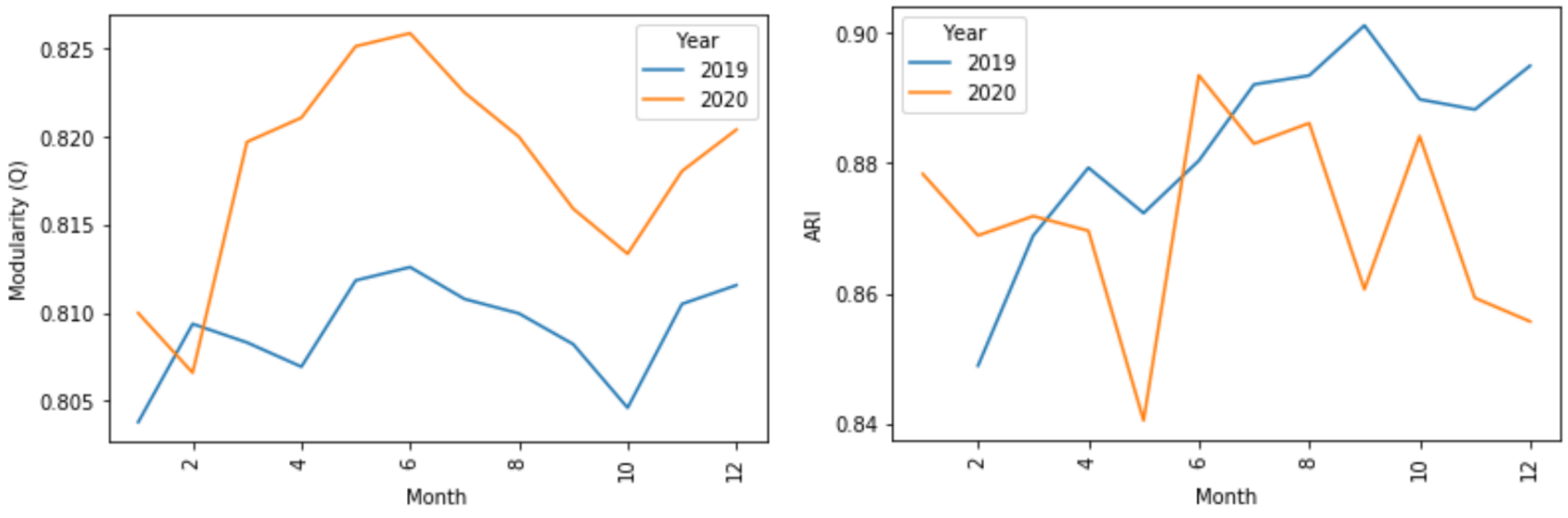

**Figure 9: Monthly Year-over-Year Microsoft Email Modularity and ARI, 2019 vs. 2020.**

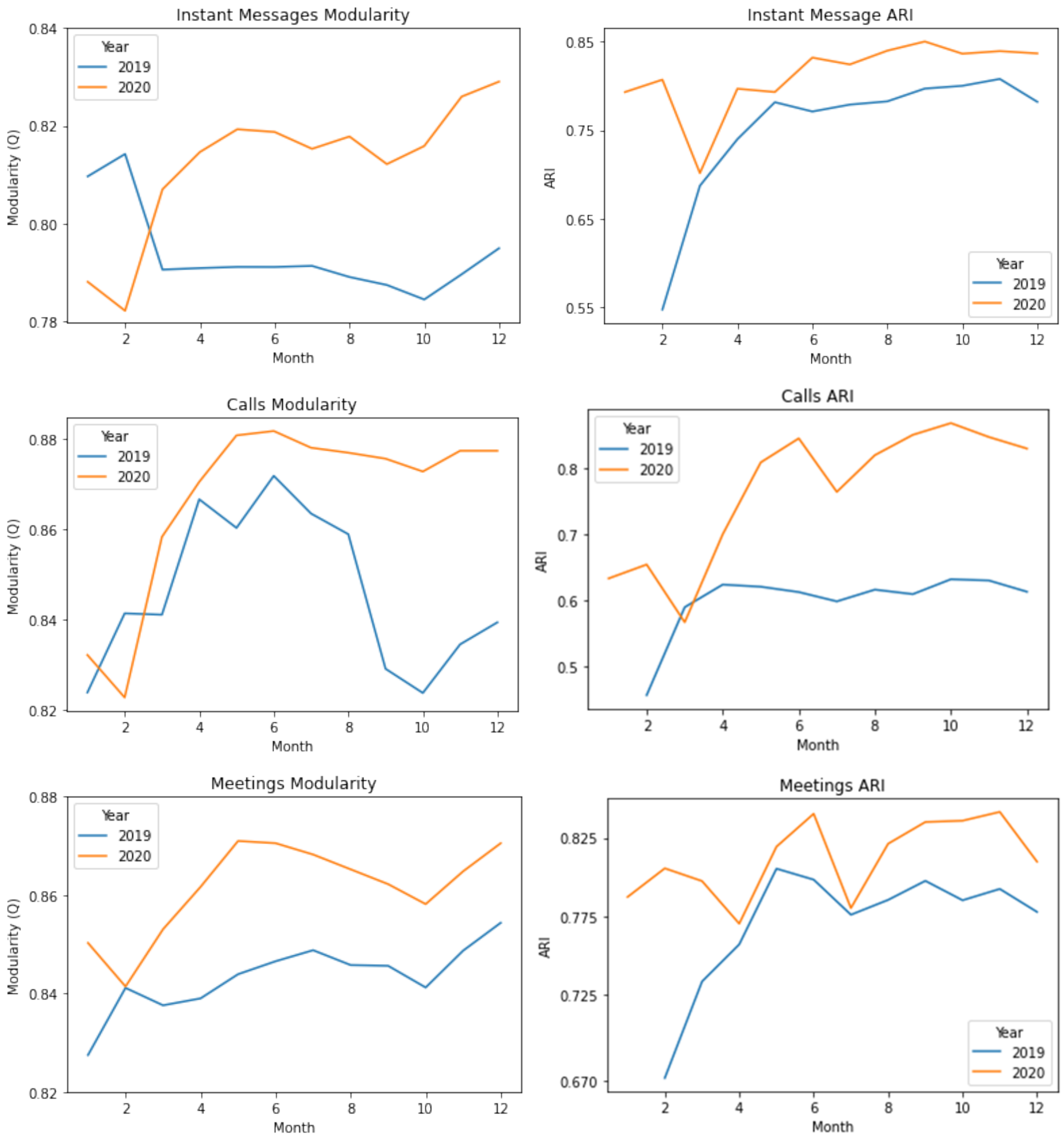

**Figure 10: Monthly Year-over-Year Microsoft Instant Message, Video Call, and Meeting Modularity and ARI, 2019 vs. 2020**.

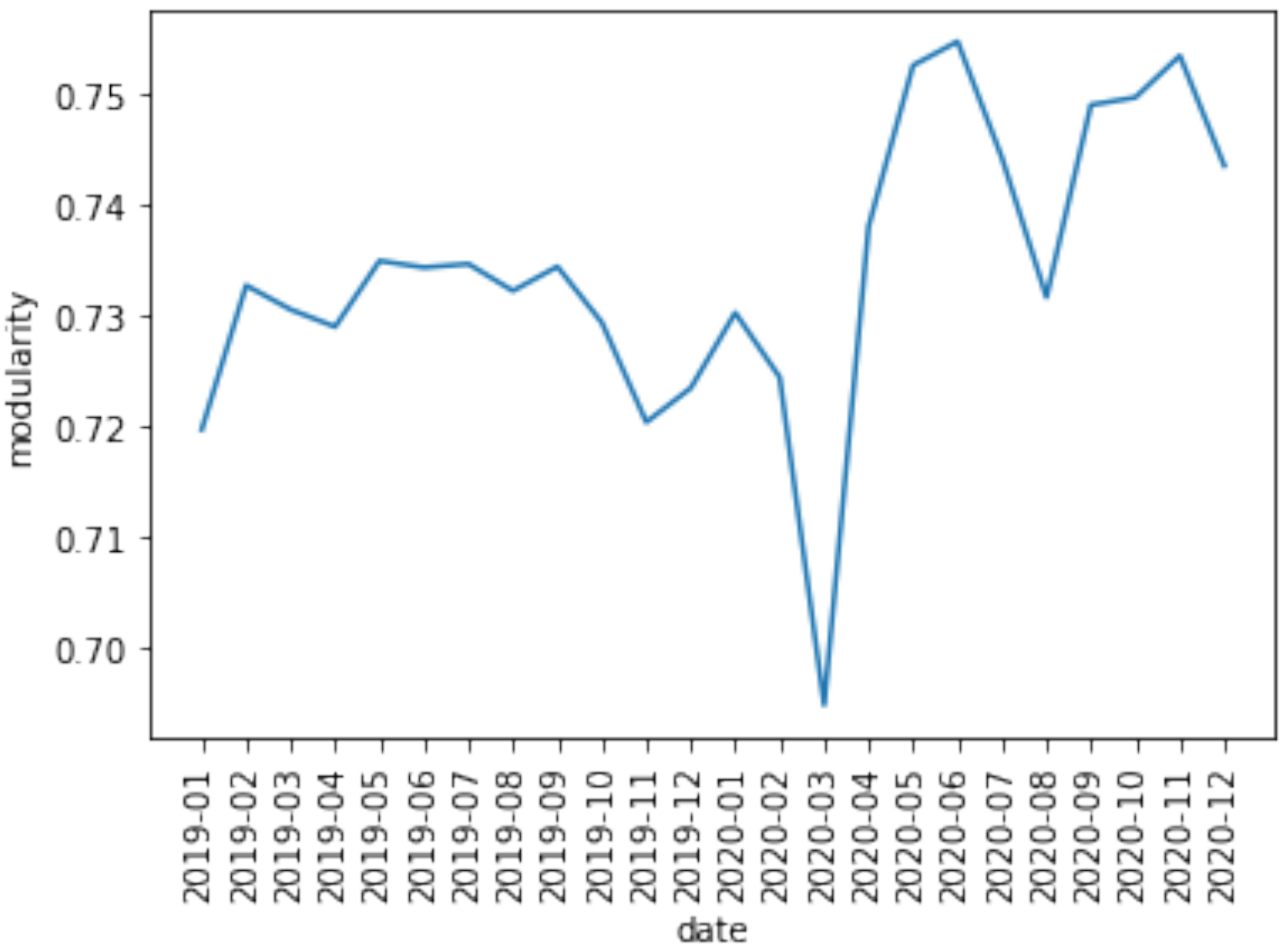

**FIGURE 11**: **Modularity for Microsoft Emails Based on Formal Organizational Structure**

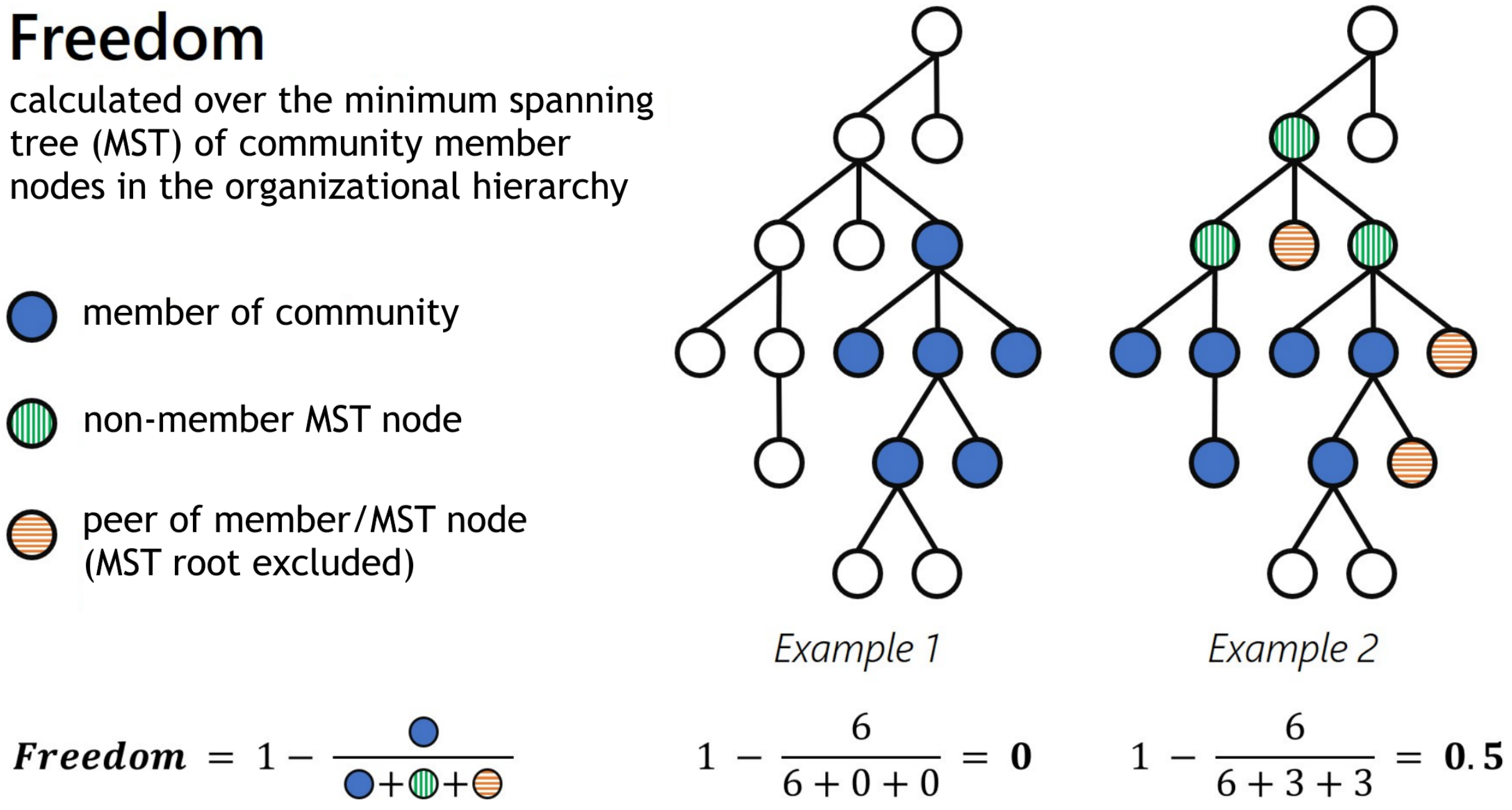

**FIGURE 12: Freedom Metric Example**

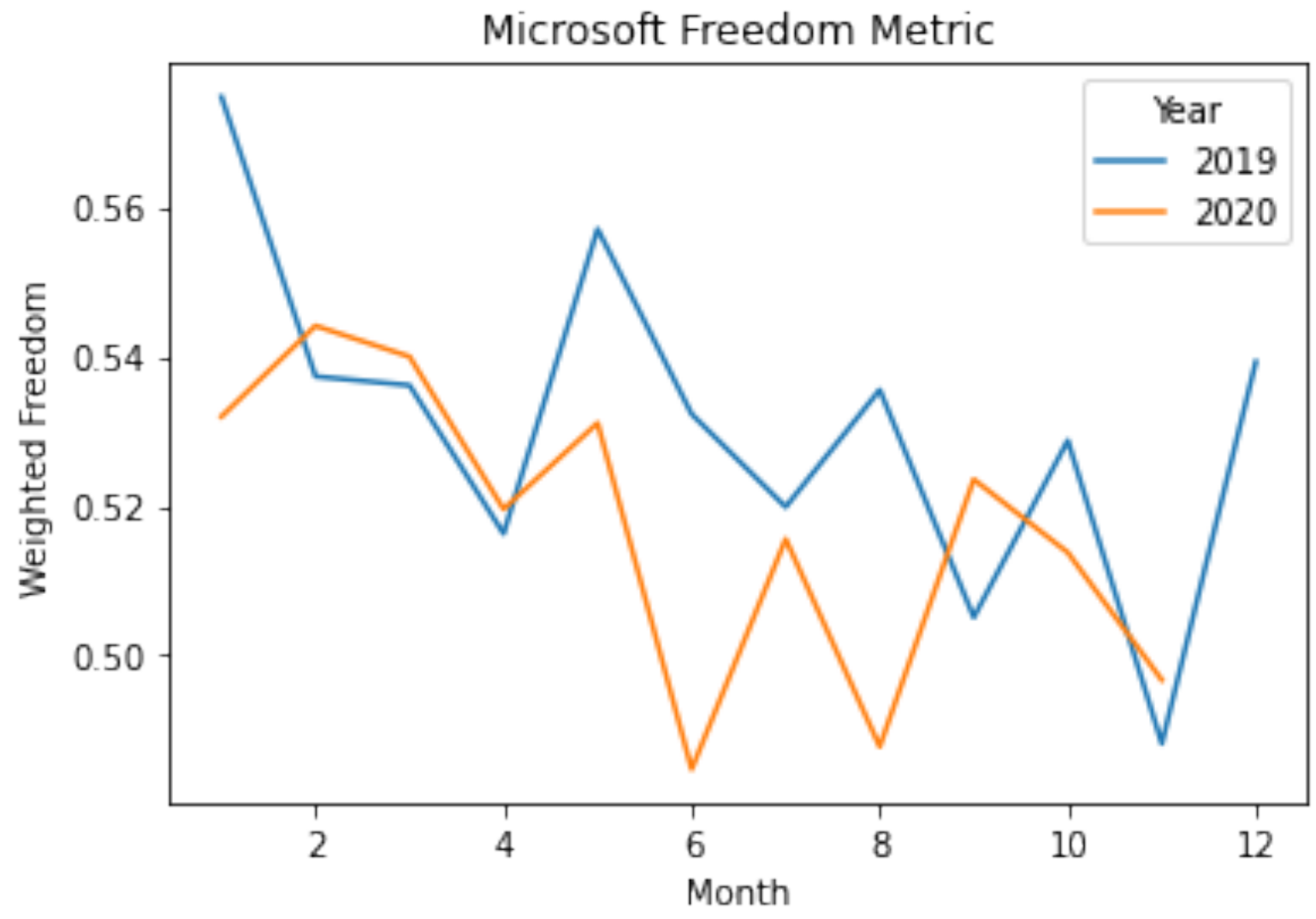

**FIGURE 13: Calculation of Freedom Metric for Microsoft Comparing Formal Organizational Structure with Leiden-derived Communities**

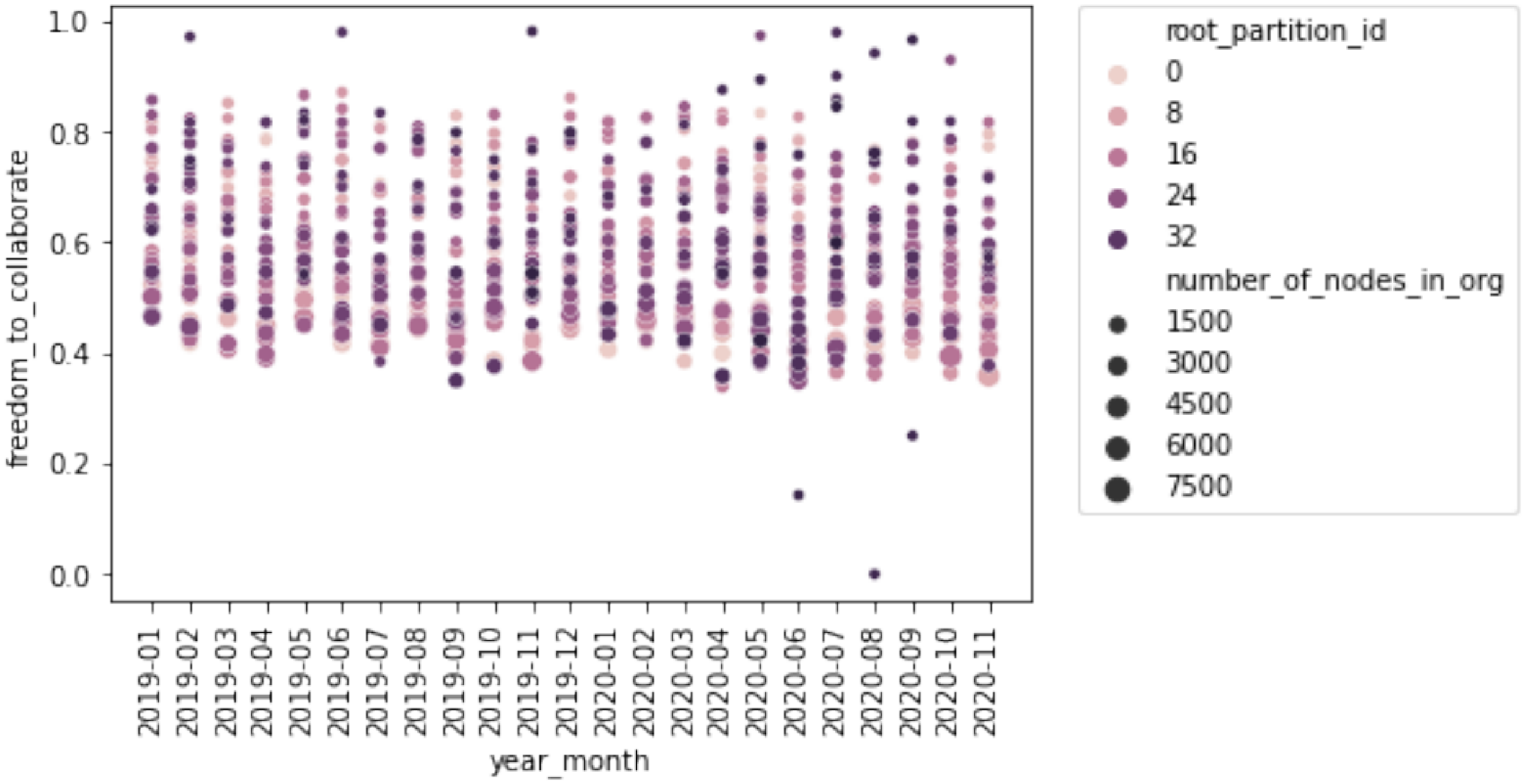

**Figure 14: Distribution of Freedom Metric by Communities at Microsoft**

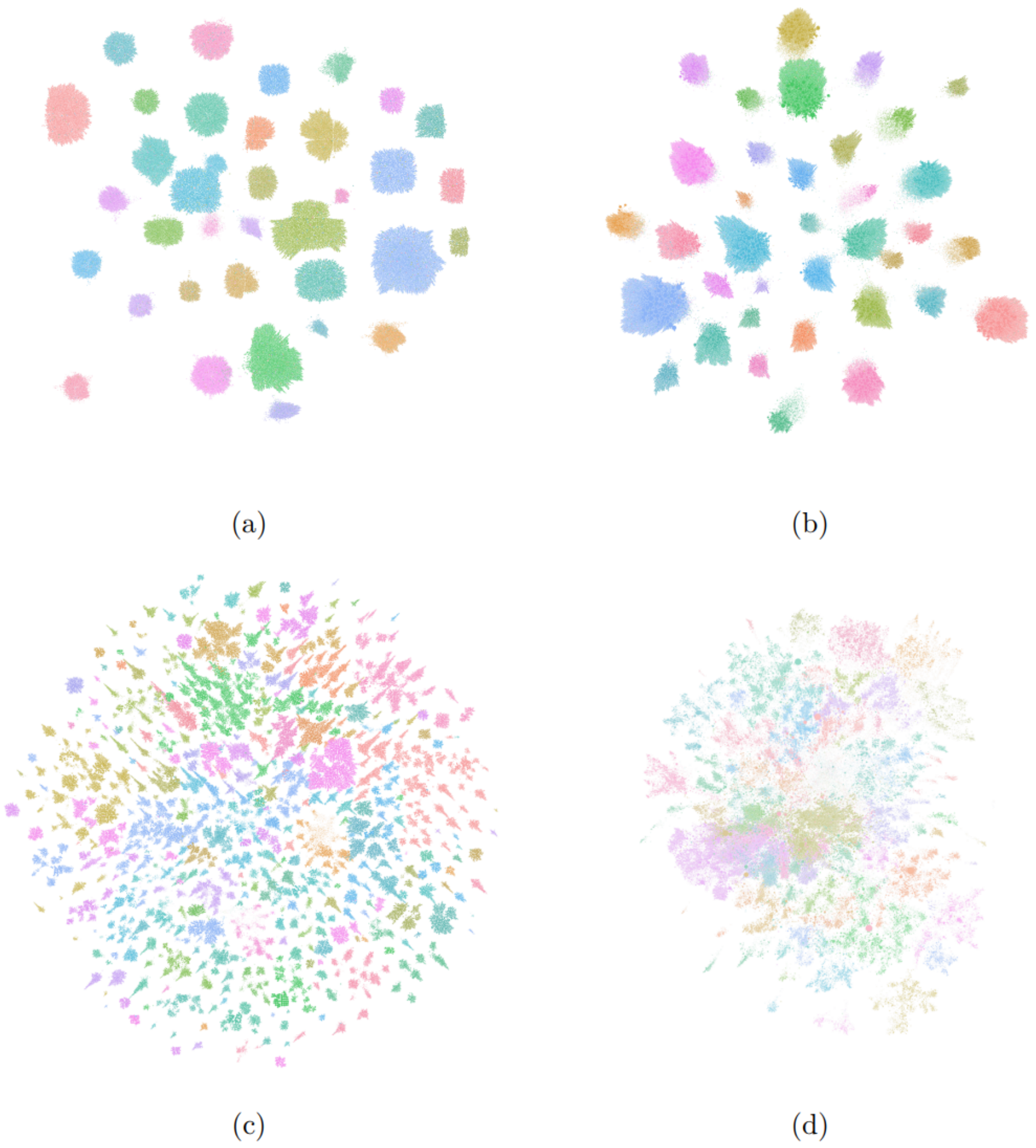

**Figure 15: Illustrative Representations of Generative Models of Microsoft, March 2020.**

**Appendix: Calculation of Modularity and ARI**

Building on Clauset et al (2004) and Traag et al (2011) given a graph $G$ with $c$ communities, we define a modularity $Q(G)$ as:

$$Q(G) = \frac{1}{2|E|}\sum_{uv}\left[A_{uv} - \frac{d_u d_v}{2|E|}\right]\delta(c_u, c_v) = \frac{1}{2|E|}\sum_{uv}\left(A_{u,v}\right)\delta(c_u, c_v) - \sum_{uv} d_u\, d_v \delta(c_u, c_v)$$
$$= \sum_{c} \frac{|E(c)|}{|E|} - \left(\frac{\sum_{v\in c} d_v}{2|E|}\right)^2,$$

where $|E|$ is the number of edges in $G$, $d_v$ is a degree of vertex $v$, $\delta(c_u, c_v)$ is 1 when $u$ and $v$ are in the same community and 0 otherwise. To provide further intuition for the calculation of modularity, consider a toy example

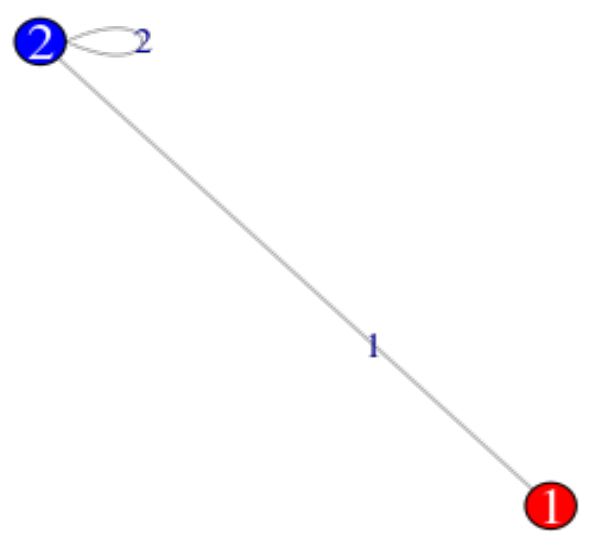

The adjacency matrix $A$ looks like this:

```
#   1 2
# 1 0 1
# 2 1 2
```

with normalization factor $2|E| = \sum_{uv} A_{uv} = 4$. The vertex degree $d_{red} = A_{11} + A_{12} = 1$, and $d_{blue} = A_{21} + A_{22} = 3$.
Then, we can calculate $Q(G)$ via

- {red,red}: $\frac{1}{2|E|}\left(A_{11} - \frac{d_{red}d_{red}}{2|E|}\right) = \frac{1}{4}(0 - 1/4) = -0.0625.$
- {blue,blue}: $\frac{1}{2|E|}\left(A_{22} - \frac{d_{blue}d_{blue}}{2|E|}\right) = \frac{1}{4}(2 - 9/4) = -0.0625.$

Summing up we get -0.0625 + -0.0625 = -0.125, which agrees with the result from `R`'s built-in function (Csardi and Nepusz, 2006) `igraph::modularity(g, membership)` = -0.125.

To calculate the ARI for the toy example above, we follow Humbert and Arabie (1985). Given a set $S$ of $n$ elements, and two groupings or partitions (e.g. clusterings) of these elements, namely $X = \{X_1, X_2, \ldots, X_r\}$ and $Y = \{Y_1, Y_2, \ldots, Y_s\}$, the overlap between $X$ and $Y$ can be summarized in a contingency table $[n_{ij}]$ where each entry $n_{ij}$ denotes the number of objects in common between $X_i$ and $Y_j$: $n_{ij} = |X_i \cap Y_j|$.

| $X \backslash Y$ | $Y_1$ | $Y_2$ | $\cdots$ | $Y_s$ | row sums |
|---|---|---|---|---|---|
| $X_1$ | $n_{11}$ | $n_{12}$ | $\cdots$ | $n_{1s}$ | $a_1$ |
| $X_2$ | $n_{21}$ | $n_{22}$ | $\cdots$ | $n_{2s}$ | $a_2$ |
| $\vdots$ | $\vdots$ | $\vdots$ | $\ddots$ | $\vdots$ | $\vdots$ |
| $X_r$ | $n_{r1}$ | $n_{r2}$ | $\cdots$ | $n_{rs}$ | $a_r$ |
| column sums | $b_1$ | $b_2$ | $\cdots$ | $b_s$ | |

The original Adjusted Rand Index using the Permutation Model is

$$ARI = \frac{\sum_{i,j}\binom{n_{i,j}}{2} - \left[\sum_i \binom{a_i}{2} \sum_j \binom{b_j}{2}\right] / \binom{n}{2}}{\frac{1}{2}\left[\sum_i \binom{a_i}{2} + \sum_j \binom{b_j}{2}\right] - \left[\sum_i \binom{a_i}{2} \sum_j \binom{b_j}{2}\right] / \binom{n}{2}}$$

where $n_{i,j}$, $a_i$, $b_j$ are values from the contingency table.

Let $x = \{$ 2, 3, 2, 2, 1, 3, 2, 2, 3, 2, 3, 2 $\}$ and $y = \{$ 2, 1, 3, 2, 2, 1, 2, 3, 1, 2, 1, 2 $\}$, then the contigency table looks like this:

```
#     y
# x   1 2 3
#   1 0 1 0
#   2 0 5 2
#   3 4 0 0
```

Breaking into components:

- $\sum_{i,j}\binom{n_{ij}}{2} = \binom{4}{2} + \binom{5}{2} + \binom{2}{2}$ = 17,
- $\sum_i \binom{a_i}{2} = \binom{7}{2} + \binom{4}{2}$ = 27,
- $\sum_j \binom{b_i}{2} = \binom{4}{2} + \binom{6}{2} + \binom{2}{2}$ = 22,
- $\binom{n}{2} = \binom{12}{2}$ = 66.

So then,

$$ARI = \frac{17 - 27 \times 22/66}{(27 + 22)/2 - 27 \times 22/66} = 0.516129,$$

and `R`'s built-in ARI function (Scrucca et al., 2016) yields the same value: `mclust::adjustedRandIndex(x,y)` = 0.516129.

**Detailed Modularity and ARI Calculations for Figure 1**

The adjacency matrix $A$ and the clustering membership $\hat{Y}$are shown below.

```
#     1 2 3 4 5 6 7 8 9 10 11 12
# 1   0 0 1 1 1 0 1 0 0  0  0  0
# 2   0 0 0 1 1 0 0 0 0  0  0  0
# 3   1 0 0 0 0 1 0 0 0  0  0  0
# 4   1 1 0 0 0 1 0 1 0  1  0  0
# 5   1 1 0 0 0 1 0 0 0  0  0  0
# 6   0 0 1 1 1 0 1 0 0  0  0  0
# 7   1 0 0 0 0 1 0 1 1  1  1  0
# 8   0 0 0 1 0 0 1 0 0  1  0  0
# 9   0 0 0 0 0 0 1 0 0  0  0  1
# 10  0 0 0 1 0 0 1 1 0  0  0  1
# 11  0 0 0 0 0 0 1 0 0  0  0  1
# 12  0 0 0 0 0 0 0 0 1  1  1  0

#  [1] 1 1 1 1 1 1 2 2 2 2 2 2
```

Then, the normalization factor $2|E| = \sum_{uv} A_{uv} = 40$. The vertex degree $d_{c_1} = 20$, and $d_{c_2} = 20$. We can calculate $Q(G_1)$ via

- $\{c_1, c_1\}$: $\frac{1}{40}(16 - 400/40) = 0.15$.
- $\{c_2, c_2\}$: $\frac{1}{40}(16 - 400/40) = 0.15$.

Summing up we get 0.15 + 0.15 = 0.3, which agrees with the result from `R`'s built-in function (Csardi and Nepuszi, 2006) `igraph::modularity(g1, membership)` = 0.3.

Because vertex labels/colors of $G_1$, $Y$ are the same as the clustering memberships, $\hat{Y}$, the calculation of $ARI(Y_1, \hat{Y})$ is very simple, that is,

```
#        memb1
#         1 2
#   blue  0 6
#   red   6 0
```

Breaking into components:

- $\sum_{i,j} \binom{n_{ij}}{2} = \binom{6}{2} + \binom{6}{2}$ = 30,
- $\sum_i \binom{a_i}{2} = \binom{6}{2}$ = 30,
- $\sum_j \binom{b_i}{2} = \binom{6}{2}$ = 30,
- $\binom{n}{2} = \binom{12}{2}$ = 66.

So then,

$$ARI = \frac{30 - 30 \times 30/66}{(30 + 30)/2 - 30 \times 30/66} = 1,$$

and `R`'s built-in ARI function (Scrucca et al., 2016) yields the same value:
`mclust::adjustedRandIndex(Y,Yhat)` = 1.

Similarly, we can calculate $Q(G_2)$ and $ARI(Y_2, \hat{Y})$ as follows. The adjacency matrix $A$ and the clustering membership $\hat{Y}$ are shown below.

```
#     1 2 3 4 5 6 7 8 9 10 11 12
# 1   0 0 0 0 0 0 1 0 0  1  1  0
# 2   0 0 0 0 0 1 0 0 0  0  0  1
# 3   0 0 0 1 0 1 0 1 0  0  0  1
# 4   0 0 1 0 0 0 0 0 1  0  0  0
# 5   0 0 0 0 0 0 0 0 0  0  1  0
# 6   0 1 1 0 0 0 0 0 1  0  0  1
# 7   1 0 0 0 0 0 0 1 0  0  1  0
# 8   0 0 1 0 0 0 1 0 0  1  1  0
# 9   0 0 0 1 0 1 0 0 0  0  0  1
# 10  1 0 0 0 0 0 0 1 0  0  1  1
# 11  1 0 0 0 1 0 1 1 0  1  0  0
# 12  0 1 1 0 0 1 0 0 1  1  0  0

#  [1] 1 2 2 2 1 2 1 1 2 1 1 2
```

Then, the normalization factor $2|E| = \sum_{uv} A_{uv} = 40$. The vertex degree $d_{c_1} = 20$, and $d_{c_2} = 20$. We can calculate $Q(G_2)$ via

- $\{c_1, c_1\}$: $\frac{1}{40}(18 - 400/40) = 0.2$.
- $\{c_2, c_2\}$: $\frac{1}{40}(18 - 400/40) = 0.2$.

Summing up we get 0.2 + 0.2 = 0.4, which agrees with the result from `R`'s built-in function `igraph::modularity(g2, membership)` = 0.4.

Now, let's calculate $ARI(Y_1, \hat{Y})$.

```
#        memb2
#          1 2
#   blue   2 4
#   red    4 2
```

Breaking into components:

- $\sum_{i,j} \binom{n_{ij}}{2} = \binom{2}{2} + \binom{4}{2} + \binom{4}{2} + \binom{2}{2}$ = 14,
- $\sum_i \binom{a_i}{2} = \binom{6}{2}$ = 30,
- $\sum_j \binom{b_i}{2} = \binom{6}{2}$ = 30,
- $\binom{n}{2} = \binom{12}{2}$ = 66.

So then,

$$ARI = \frac{14 - 30 \times 30/66}{(30 + 30)/2 - 30 \times 30/66} = 0.0222222,$$

and `R`'s built-in ARI function yields the same value: `mclust::adjustedRandIndex(Y,Yhat)` = 0.0222222. The summary table is shown below:

| graph | Q | Khat | ARI |
|---|---|---|---|
| G1 | 0.3 | 2 | 1.00 |
| G2 | 0.4 | 2 | 0.02 |